\PassOptionsToPackage{round,authoryear}{natbib}
\PassOptionsToPackage{table,dvipsnames}{xcolor}
\documentclass[]{style_company_light}

\usepackage[utf8]{inputenc}
\usepackage{CJKutf8}
\usepackage{hyperref}
\usepackage{url}
\usepackage{booktabs}
\usepackage{amsfonts}
\usepackage{amsmath}
\usepackage{amsthm}
\usepackage{microtype}
\usepackage{graphicx}
\usepackage{multirow}
\usepackage{enumitem}
\usepackage{adjustbox}
\usepackage{array}
\usepackage{tabularx}
\usepackage{longtable}
\usepackage{amssymb}
\usepackage{multicol}
\usepackage{xcolor}
\usepackage{xspace}
\usepackage{tikz}
\usetikzlibrary{arrows.meta}
\usepackage{pgfplots}
\pgfplotsset{compat=1.18}
\usepackage{algorithm}
\usepackage{algpseudocode}
\floatname{algorithm}{Algorithm}
\newtheorem{definition}{Definition}
\newenvironment{figgroup}[2]{%
  \begin{adjustbox}{max totalsize={\linewidth}{0.90\textheight},center}%
  \begin{minipage}{\linewidth}%
  \begin{tcolorbox}[enhanced,colback=white,colframe=#1!60,boxrule=0pt,frame hidden,arc=0pt,
    left=4pt,right=4pt,top=4pt,bottom=4pt,halign=center,
    title={\sffamily\bfseries #2},coltitle=#1,colbacktitle=white,fonttitle=\large,
    attach title to upper,after title={\par\smallskip}]}%
  {\end{tcolorbox}\end{minipage}\end{adjustbox}}
\newcommand{\goldmedal}{%
  \tikz[baseline=-0.5ex,x=1pt,y=1pt]{%
    \fill[red!65!black] (-1.8,3.2)--(-0.9,0.8)--(0.6,1.5)--(-0.3,3.6)--cycle;
    \fill[red!50!black] (1.8,3.2)--(0.9,0.8)--(-0.6,1.5)--(0.3,3.6)--cycle;
    \filldraw[fill=yellow!75!orange,draw=orange!75!black,line width=0.25pt]
      (0,0) circle (2pt);
    \fill[orange!60!black] (0,1.1)--(0.25,0.3)--(1.05,0.3)--(0.4,-0.2)
      --(0.65,-1)--(0,-0.5)--(-0.65,-1)--(-0.4,-0.2)--(-1.05,0.3)
      --(-0.25,0.3)--cycle;}}
\newcommand{\goldrank}{\rlap{\kern0.15em\goldmedal}}
\definecolor{grpsuccess}{HTML}{25877C}
\definecolor{grpprogress}{HTML}{2465A5}
\definecolor{grpstagnation}{HTML}{C56049}
\definecolor{grpcontinuation}{HTML}{6A7689}
\definecolor{grpoffpolicy}{HTML}{7255A3}
\ifdefined\pdfsuppresswarningpagegroup\pdfsuppresswarningpagegroup=1\fi

\newtoggle{hldraft}
\toggletrue{hldraft}

\newcommand{\hlnumtbdlist}{}

\newcommand{\defval}[2]{\csdef{hlnum@#1}{#2}\csundef{hlntbd@#1}}
\newcommand{\defTBD}[2]{%
  \csdef{hlnum@#1}{#2}\csdef{hlntbd@#1}{}\listadd{\hlnumtbdlist}{#1}}

\newcommand{\val}[1]{%
  \ifcsdef{hlnum@#1}{%
    \ifcsdef{hlntbd@#1}{%
      \iftoggle{hldraft}{\textcolor{brandaccent}{\bfseries\csuse{hlnum@#1}}}%
        {\PackageError{hlnumbers}{Unmeasured key '#1' used with hldraft off}%
          {Key '#1' is still \string\defTBD. Measure it and convert it to
           \string\defval, or re-enable \string\toggletrue{hldraft}. The
           camera-ready build will not proceed with placeholders in it.}%
         \csuse{hlnum@#1}}%
    }{\csuse{hlnum@#1}}%
  }{%
    \textcolor{red}{\bfseries[?\,\texttt{#1}]}%
    \PackageWarning{hlnumbers}{Undefined number key '#1'}%
  }%
}

\newcommand{\hlnumtbdlog}[1]{\typeout{  #1 = \csuse{hlnum@#1}}}
\AtEndDocument{%
  \iftoggle{hldraft}{%
    \typeout{^^J===== HL OUTSTANDING MEASUREMENTS (draft mode) =====}%
    \forlistloop{\hlnumtbdlog}{\hlnumtbdlist}%
    \typeout{===== end =====^^J}%
  }{}}

\defval{ngames}{12}
\defval{newgames}{4}
\defval{basegames}{8}
\defval{completedmodels}{7}
\defval{bestconfigrankonegames}{6}
\defval{unconqueredgames}{6}
\defval{mainruns}{60}
\defval{smallbudget}{128}
\defval{largebudget}{16}
\defval{extendedsmall}{384}
\defval{extendedlarge}{48}
\defval{clonebudget}{32}
\defval{clonetrials}{48}
\defval{clonerankone}{5}
\defval{cloneranktwo}{15}
\defval{clonerankthree}{25}
\defval{clonerankfour}{35}
\defval{ladderstart}{25}
\defval{ladderstep}{0, 1, or 2}
\defval{toplimit}{5}
\defval{batchmax}{8}
\defval{referenceRank}{40}
\defval{continuedcomplete}{4}
\defval{continuedimprove}{2}
\defval{continuedrankimprove}{2}
\defval{replaycomplete}{7}
\defval{replaybetter}{4}
\defval{replayworse}{3}
\defval{snapshotdate}{September 17, 2026}
\defval{snapshotwindow}{22:45:13--22:45:35 CST (UTC+8)}
\defval{elobase}{10}
\defval{eloscale}{400}
\defval{antgridsize}{19}
\defval{antbasehp}{50}
\defval{antweapons}{4}
\defval{main-rollman-deepseek-v4-pro-elo}{581.169}
\defval{main-rollman-deepseek-v4-pro-rank}{1}
\defval{main-rollman-kimi-k3-elo}{534.174}
\defval{main-rollman-kimi-k3-rank}{1}
\defval{main-rollman-longcat-2.0-elo}{54.038}
\defval{main-rollman-longcat-2.0-rank}{23}
\defval{main-rollman-glm-5.3-elo}{581.169}
\defval{main-rollman-glm-5.3-rank}{1}
\defval{main-pacman-deepseek-v4-pro-elo}{1710.120}
\defval{main-pacman-deepseek-v4-pro-rank}{11}
\defval{main-pacman-kimi-k3-elo}{1941.570}
\defval{main-pacman-kimi-k3-rank}{8}
\defval{main-pacman-longcat-2.0-elo}{1796.697}
\defval{main-pacman-longcat-2.0-rank}{10}
\defval{main-pacman-glm-5.3-elo}{1726.478}
\defval{main-pacman-glm-5.3-rank}{10}
\defval{main-antwar-deepseek-v4-pro-elo}{1498.113}
\defval{main-antwar-deepseek-v4-pro-rank}{5}
\defval{main-antwar-kimi-k3-elo}{1075.921}
\defval{main-antwar-kimi-k3-rank}{11}
\defval{main-antwar-longcat-2.0-elo}{807.502}
\defval{main-antwar-longcat-2.0-rank}{33}
\defval{main-antwar-glm-5.3-elo}{1208.652}
\defval{main-antwar-glm-5.3-rank}{8}
\defval{main-snakego-deepseek-v4-pro-elo}{1192.663}
\defval{main-snakego-deepseek-v4-pro-rank}{3}
\defval{main-snakego-kimi-k3-elo}{1165.873}
\defval{main-snakego-kimi-k3-rank}{3}
\defval{main-snakego-longcat-2.0-elo}{436.375}
\defval{main-snakego-longcat-2.0-rank}{71}
\defval{main-snakego-glm-5.3-elo}{659.256}
\defval{main-snakego-glm-5.3-rank}{44}
\defval{main-aquawar-deepseek-v4-pro-elo}{1417.351}
\defval{main-aquawar-deepseek-v4-pro-rank}{2}
\defval{main-aquawar-kimi-k3-elo}{1429.469}
\defval{main-aquawar-kimi-k3-rank}{2}
\defval{main-aquawar-longcat-2.0-elo}{943.552}
\defval{main-aquawar-longcat-2.0-rank}{68}
\defval{main-aquawar-glm-5.3-elo}{1450.865}
\defval{main-aquawar-glm-5.3-rank}{1}
\defval{main-generals-deepseek-v4-pro-elo}{1367.495}
\defval{main-generals-deepseek-v4-pro-rank}{22}
\defval{main-generals-kimi-k3-elo}{1381.374}
\defval{main-generals-kimi-k3-rank}{20}
\defval{main-generals-longcat-2.0-elo}{1203.061}
\defval{main-generals-longcat-2.0-rank}{35}
\defval{main-generals-glm-5.3-elo}{1388.460}
\defval{main-generals-glm-5.3-rank}{20}
\defval{main-lostspace-deepseek-v4-pro-elo}{1865.323}
\defval{main-lostspace-deepseek-v4-pro-rank}{3}
\defval{main-lostspace-kimi-k3-elo}{1763.789}
\defval{main-lostspace-kimi-k3-rank}{3}
\defval{main-lostspace-longcat-2.0-elo}{1350.185}
\defval{main-lostspace-longcat-2.0-rank}{46}
\defval{main-lostspace-glm-5.3-elo}{1625.213}
\defval{main-lostspace-glm-5.3-rank}{7}
\defval{main-miracle-deepseek-v4-pro-elo}{1591.052}
\defval{main-miracle-deepseek-v4-pro-rank}{11}
\defval{main-miracle-kimi-k3-elo}{1602.626}
\defval{main-miracle-kimi-k3-rank}{11}
\defval{main-miracle-longcat-2.0-elo}{918.751}
\defval{main-miracle-longcat-2.0-rank}{126}
\defval{main-miracle-glm-5.3-elo}{1436.775}
\defval{main-miracle-glm-5.3-rank}{26}
\defval{main-antwar2-deepseek-v4-pro-elo}{1432.628}
\defval{main-antwar2-deepseek-v4-pro-rank}{34}
\defval{main-antwar2-kimi-k3-elo}{1908.547}
\defval{main-antwar2-kimi-k3-rank}{12}
\defval{main-antwar2-longcat-2.0-elo}{1432.628}
\defval{main-antwar2-longcat-2.0-rank}{34}
\defval{main-antwar2-glm-5.3-elo}{1432.628}
\defval{main-antwar2-glm-5.3-rank}{34}
\defval{main-monecraft-deepseek-v4-pro-elo}{2248.316}
\defval{main-monecraft-deepseek-v4-pro-rank}{4}
\defval{main-monecraft-kimi-k3-elo}{2229.945}
\defval{main-monecraft-kimi-k3-rank}{4}
\defval{main-monecraft-longcat-2.0-elo}{2164.321}
\defval{main-monecraft-longcat-2.0-rank}{8}
\defval{main-monecraft-glm-5.3-elo}{2179.720}
\defval{main-monecraft-glm-5.3-rank}{6}
\defval{main-lota-deepseek-v4-pro-elo}{2119.279}
\defval{main-lota-deepseek-v4-pro-rank}{22}
\defval{main-lota-kimi-k3-elo}{2105.679}
\defval{main-lota-kimi-k3-rank}{24}
\defval{main-lota-longcat-2.0-elo}{1556.673}
\defval{main-lota-longcat-2.0-rank}{81}
\defval{main-lota-glm-5.3-elo}{1814.760}
\defval{main-lota-glm-5.3-rank}{48}
\defval{main-dorado-deepseek-v4-pro-elo}{1811.432}
\defval{main-dorado-deepseek-v4-pro-rank}{34}
\defval{main-dorado-kimi-k3-elo}{1922.817}
\defval{main-dorado-kimi-k3-rank}{22}
\defval{main-dorado-longcat-2.0-elo}{1503.163}
\defval{main-dorado-longcat-2.0-rank}{165}
\defval{main-dorado-glm-5.3-elo}{2023.352}
\defval{main-dorado-glm-5.3-rank}{16}
\defval{cont-base-rollman-elo}{581.169}
\defval{cont-base-rollman-rank}{1}
\defval{cont-rollman-elo}{636.858}
\defval{cont-rollman-rank}{1}
\defval{cont-delta-rollman}{+55.689}
\defval{cont-small-rollman}{124}
\defval{cont-large-rollman}{16/16}
\defval{cont-base-pacman-elo}{1726.478}
\defval{cont-base-pacman-rank}{10}
\defval{cont-pacman-elo}{1778.376}
\defval{cont-pacman-rank}{10}
\defval{cont-delta-pacman}{+51.898}
\defval{cont-small-pacman}{128}
\defval{cont-large-pacman}{16/16}
\defval{cont-base-antwar-elo}{1208.652}
\defval{cont-base-antwar-rank}{8}
\defval{cont-antwar-elo}{1239.481}
\defval{cont-antwar-rank}{8}
\defval{cont-delta-antwar}{+30.829}
\defval{cont-small-antwar}{128}
\defval{cont-large-antwar}{16/16}
\defval{cont-base-snakego-elo}{659.256}
\defval{cont-base-snakego-rank}{44}
\defval{cont-snakego-elo}{697.859}
\defval{cont-snakego-rank}{40}
\defval{cont-delta-snakego}{+38.603}
\defval{cont-small-snakego}{128}
\defval{cont-large-snakego}{11/11}
\defval{cont-base-aquawar-elo}{1450.865}
\defval{cont-base-aquawar-rank}{1}
\defval{cont-aquawar-elo}{1533.509}
\defval{cont-aquawar-rank}{1}
\defval{cont-delta-aquawar}{+82.644}
\defval{cont-small-aquawar}{128}
\defval{cont-large-aquawar}{16/16}
\defval{cont-base-generals-elo}{1388.460}
\defval{cont-base-generals-rank}{20}
\defval{cont-generals-elo}{1417.866}
\defval{cont-generals-rank}{18}
\defval{cont-delta-generals}{+29.406}
\defval{cont-small-generals}{128}
\defval{cont-large-generals}{16/16}
\defval{cont-base-lostspace-elo}{1625.213}
\defval{cont-base-lostspace-rank}{7}
\defval{cont-lostspace-elo}{1632.538}
\defval{cont-lostspace-rank}{7}
\defval{cont-delta-lostspace}{+7.325}
\defval{cont-small-lostspace}{128}
\defval{cont-large-lostspace}{16/16}
\defval{cont-base-miracle-elo}{1436.775}
\defval{cont-base-miracle-rank}{26}
\defval{cont-miracle-elo}{1436.775}
\defval{cont-miracle-rank}{26}
\defval{cont-delta-miracle}{+0.000}
\defval{cont-small-miracle}{112}
\defval{cont-large-miracle}{16/16}
\defval{cont-base-antwar2-elo}{1432.628}
\defval{cont-base-antwar2-rank}{34}
\defval{cont-antwar2-elo}{1432.628}
\defval{cont-antwar2-rank}{34}
\defval{cont-delta-antwar2}{+0.000}
\defval{cont-small-antwar2}{24}
\defval{cont-large-antwar2}{0/0}
\defval{cont-base-monecraft-elo}{2179.720}
\defval{cont-base-monecraft-rank}{6}
\defval{cont-monecraft-elo}{2179.720}
\defval{cont-monecraft-rank}{6}
\defval{cont-delta-monecraft}{+0.000}
\defval{cont-small-monecraft}{56}
\defval{cont-large-monecraft}{16/16}
\defval{cont-base-lota-elo}{1814.760}
\defval{cont-base-lota-rank}{48}
\defval{cont-lota-elo}{1836.253}
\defval{cont-lota-rank}{46}
\defval{cont-delta-lota}{+21.493}
\defval{cont-small-lota}{80}
\defval{cont-large-lota}{16/16}
\defval{cont-base-dorado-elo}{2023.352}
\defval{cont-base-dorado-rank}{16}
\defval{cont-dorado-elo}{2023.352}
\defval{cont-dorado-rank}{16}
\defval{cont-delta-dorado}{+0.000}
\defval{cont-small-dorado}{128}
\defval{cont-large-dorado}{2/3}
\defval{opp-base-rollman-elo}{581.169}
\defval{opp-base-rollman-rank}{1}
\defval{opp-rollman-ladder-elo}{581.169}
\defval{opp-rollman-ladder-rank}{1}
\defval{opp-rollman-random-elo}{455.800}
\defval{opp-rollman-random-rank}{5}
\defval{opp-rollman-top5-elo}{534.174}
\defval{opp-rollman-top5-rank}{1}
\defval{opp-base-pacman-elo}{1726.478}
\defval{opp-base-pacman-rank}{10}
\defval{opp-pacman-ladder-elo}{1778.376}
\defval{opp-pacman-ladder-rank}{10}
\defval{opp-pacman-random-elo}{2014.368}
\defval{opp-pacman-random-rank}{6}
\defval{opp-pacman-top5-elo}{2014.368}
\defval{opp-pacman-top5-rank}{6}
\defval{opp-base-antwar-elo}{1208.652}
\defval{opp-base-antwar-rank}{8}
\defval{opp-antwar-ladder-elo}{1154.428}
\defval{opp-antwar-ladder-rank}{9}
\defval{opp-antwar-random-elo}{1223.703}
\defval{opp-antwar-random-rank}{8}
\defval{opp-antwar-top5-elo}{1107.540}
\defval{opp-antwar-top5-rank}{10}
\defval{opp-base-snakego-elo}{659.256}
\defval{opp-base-snakego-rank}{44}
\defval{opp-snakego-ladder-elo}{1192.663}
\defval{opp-snakego-ladder-rank}{3}
\defval{opp-snakego-random-elo}{1273.228}
\defval{opp-snakego-random-rank}{1}
\defval{opp-snakego-top5-elo}{1153.313}
\defval{opp-snakego-top5-rank}{4}
\defval{opp-base-aquawar-elo}{1450.865}
\defval{opp-base-aquawar-rank}{1}
\defval{opp-aquawar-ladder-elo}{1425.374}
\defval{opp-aquawar-ladder-rank}{2}
\defval{opp-aquawar-random-elo}{1450.865}
\defval{opp-aquawar-random-rank}{1}
\defval{opp-aquawar-top5-elo}{1433.623}
\defval{opp-aquawar-top5-rank}{2}
\defval{opp-base-generals-elo}{1388.460}
\defval{opp-base-generals-rank}{20}
\defval{opp-generals-ladder-elo}{1223.929}
\defval{opp-generals-ladder-rank}{33}
\defval{opp-generals-random-elo}{1279.405}
\defval{opp-generals-random-rank}{27}
\defval{opp-generals-top5-elo}{1367.495}
\defval{opp-generals-top5-rank}{22}
\defval{opp-base-lostspace-elo}{1625.213}
\defval{opp-base-lostspace-rank}{7}
\defval{opp-lostspace-ladder-elo}{1610.916}
\defval{opp-lostspace-ladder-rank}{10}
\defval{opp-lostspace-random-elo}{1687.878}
\defval{opp-lostspace-random-rank}{5}
\defval{opp-lostspace-top5-elo}{1785.988}
\defval{opp-lostspace-top5-rank}{3}
\defval{opp-base-miracle-elo}{1436.775}
\defval{opp-base-miracle-rank}{26}
\defval{opp-miracle-ladder-elo}{1509.312}
\defval{opp-miracle-ladder-rank}{17}
\defval{opp-miracle-random-elo}{1193.736}
\defval{opp-miracle-random-rank}{74}
\defval{opp-miracle-top5-elo}{1283.574}
\defval{opp-miracle-top5-rank}{55}
\defval{opp-base-antwar2-elo}{1432.628}
\defval{opp-base-antwar2-rank}{34}
\defval{opp-base-monecraft-elo}{2179.720}
\defval{opp-base-monecraft-rank}{6}
\defval{opp-monecraft-ladder-elo}{2164.321}
\defval{opp-monecraft-ladder-rank}{8}
\defval{opp-monecraft-random-elo}{2229.945}
\defval{opp-monecraft-random-rank}{4}
\defval{opp-monecraft-top5-elo}{2267.691}
\defval{opp-monecraft-top5-rank}{3}
\defval{opp-base-lota-elo}{1814.760}
\defval{opp-base-lota-rank}{48}
\defval{opp-lota-ladder-elo}{2119.279}
\defval{opp-lota-ladder-rank}{22}
\defval{opp-lota-random-elo}{1961.921}
\defval{opp-lota-random-rank}{37}
\defval{opp-lota-top5-elo}{1528.576}
\defval{opp-lota-top5-rank}{92}
\defval{opp-base-dorado-elo}{2023.352}
\defval{opp-base-dorado-rank}{16}
\defval{opp-dorado-ladder-elo}{907.893}
\defval{opp-dorado-ladder-rank}{308}
\defval{opp-dorado-random-elo}{1683.709}
\defval{opp-dorado-random-rank}{54}
\defval{opp-dorado-top5-elo}{918.196}
\defval{opp-dorado-top5-rank}{308}
\defval{replay-base-rollman-elo}{581.169}
\defval{replay-base-rollman-rank}{1}
\defval{replay-rollman-elo}{534.174}
\defval{replay-rollman-rank}{1}
\defval{replay-base-pacman-elo}{1726.478}
\defval{replay-base-pacman-rank}{10}
\defval{replay-pacman-elo}{1815.567}
\defval{replay-pacman-rank}{10}
\defval{replay-base-antwar-elo}{1208.652}
\defval{replay-base-antwar-rank}{8}
\defval{replay-antwar-elo}{833.528}
\defval{replay-antwar-rank}{29}
\defval{replay-base-snakego-elo}{659.256}
\defval{replay-base-snakego-rank}{44}
\defval{replay-snakego-elo}{1118.365}
\defval{replay-snakego-rank}{5}
\defval{replay-base-aquawar-elo}{1450.865}
\defval{replay-base-aquawar-rank}{1}
\defval{replay-aquawar-elo}{1417.351}
\defval{replay-aquawar-rank}{2}
\defval{replay-base-generals-elo}{1388.460}
\defval{replay-base-generals-rank}{20}
\defval{replay-generals-elo}{1315.421}
\defval{replay-generals-rank}{26}
\defval{replay-base-lostspace-elo}{1625.213}
\defval{replay-base-lostspace-rank}{7}
\defval{replay-lostspace-elo}{1052.874}
\defval{replay-lostspace-rank}{77}
\defval{replay-base-miracle-elo}{1436.775}
\defval{replay-base-miracle-rank}{26}
\defval{replay-miracle-elo}{1491.192}
\defval{replay-miracle-rank}{19}
\defval{replay-base-antwar2-elo}{1432.628}
\defval{replay-base-antwar2-rank}{34}
\defval{replay-base-monecraft-elo}{2179.720}
\defval{replay-base-monecraft-rank}{6}
\defval{replay-monecraft-elo}{1919.810}
\defval{replay-monecraft-rank}{16}
\defval{replay-base-lota-elo}{1814.760}
\defval{replay-base-lota-rank}{48}
\defval{replay-lota-elo}{2076.120}
\defval{replay-lota-rank}{27}
\defval{replay-base-dorado-elo}{2023.352}
\defval{replay-base-dorado-rank}{16}
\defval{replay-dorado-elo}{913.085}
\defval{replay-dorado-rank}{308}
\defval{clone-rollman-5-tokens}{25.86}
\defval{clone-rollman-15-tokens}{6.62}
\defval{clone-rollman-25-tokens}{8.85}
\defval{clone-rollman-35-tokens}{9.47}
\defval{clone-pacman-5-tokens}{28.77}
\defval{clone-pacman-15-tokens}{10.96}
\defval{clone-pacman-25-tokens}{10.10}
\defval{clone-pacman-35-tokens}{22.99}
\defval{clone-antwar-5-tokens}{34.12}
\defval{clone-antwar-15-tokens}{20.35}
\defval{clone-antwar-25-tokens}{13.03}
\defval{clone-antwar-35-tokens}{7.34}
\defval{clone-snakego-5-tokens}{18.80}
\defval{clone-snakego-15-tokens}{20.96}
\defval{clone-snakego-25-tokens}{17.32}
\defval{clone-snakego-35-tokens}{15.38}
\defval{clone-aquawar-5-tokens}{27.18}
\defval{clone-aquawar-15-tokens}{14.75}
\defval{clone-aquawar-25-tokens}{27.07}
\defval{clone-aquawar-35-tokens}{33.30}
\defval{clone-generals-5-tokens}{24.55}
\defval{clone-generals-15-tokens}{10.93}
\defval{clone-generals-25-tokens}{12.33}
\defval{clone-generals-35-tokens}{41.18}
\defval{clone-lostspace-5-tokens}{14.51}
\defval{clone-lostspace-15-tokens}{56.34}
\defval{clone-lostspace-25-tokens}{26.53}
\defval{clone-lostspace-35-tokens}{32.26}
\defval{clone-miracle-5-tokens}{24.93}
\defval{clone-miracle-15-tokens}{29.12}
\defval{clone-miracle-25-tokens}{19.42}
\defval{clone-miracle-35-tokens}{16.39}
\defval{clone-antwar2-5-tokens}{35.35}
\defval{clone-antwar2-15-tokens}{21.08}
\defval{clone-antwar2-25-tokens}{48.39}
\defval{clone-antwar2-35-tokens}{21.62}
\defval{clone-monecraft-5-tokens}{14.12}
\defval{clone-monecraft-15-tokens}{24.95}
\defval{clone-monecraft-25-tokens}{10.34}
\defval{clone-monecraft-35-tokens}{13.85}
\defval{clone-lota-5-tokens}{19.47}
\defval{clone-lota-15-tokens}{19.82}
\defval{clone-lota-25-tokens}{23.70}
\defval{clone-lota-35-tokens}{40.14}
\defval{clone-dorado-5-tokens}{24.05}
\defval{clone-dorado-15-tokens}{17.84}
\defval{clone-dorado-25-tokens}{14.52}
\defval{clone-dorado-35-tokens}{36.74}
\defval{deepseek-v4-pro-completed}{12}
\defval{deepseek-v4-pro-first}{1}
\defval{deepseek-v4-pro-tokens}{798.30}
\defval{kimi-k3-completed}{12}
\defval{kimi-k3-first}{1}
\defval{kimi-k3-tokens}{355.11}
\defval{longcat-2.0-completed}{12}
\defval{longcat-2.0-first}{0}
\defval{longcat-2.0-tokens}{552.04}
\defval{glm-5.3-completed}{12}
\defval{glm-5.3-first}{2}
\defval{glm-5.3-tokens}{340.15}
\defval{gpt5.6sol-completed}{0}
\defval{opus5-completed}{0}
\defval{pool-rollman}{64}
\defval{pool-pacman}{44}
\defval{pool-antwar}{114}
\defval{pool-snakego}{141}
\defval{pool-aquawar}{170}
\defval{pool-generals}{197}
\defval{pool-lostspace}{111}
\defval{pool-miracle}{253}
\defval{pool-antwar2}{191}
\defval{pool-monecraft}{112}
\defval{pool-lota}{200}
\defval{pool-dorado}{323}

\newcommand{\benchname}{\mbox{AAArena}\xspace}

\title{Can AI Agents Learn Their Way to the Top?\\
Evaluating Heuristic Learning in a Long-Running Game Agent Competition}
\author{Kaisen Yang$^{1,*,\dagger}$, Qingle Liu$^{1,*,\dagger}$, Kejin Wang$^{1,*}$, Yicheng Zhao$^{1,*}$, Jieming Li$^{1,*}$, Shenghan Zheng$^{1,*}$, Ruize Yang$^{1,*}$, Bojun Yang$^{1,*}$, Heng Gong$^{1}$, Xiang Gao$^{1}$, Lanyue Zhang$^{1}$, Kaiyu Zhong$^{1}$, Zhuo Liu$^{1}$, Shaoxuan Li$^{1}$, Chengxi Li$^{1}$, Yong Yan$^{1}$,  Weixuan Zhang$^{1}$, Tianwei Luo$^{1}$, Situ Wang$^{1}$, Youjie Zheng$^{1}$, Sihan Zhao$^{1}$, Shengyuan Wang$^{1,2}$, Huan-ang Gao$^{1}$, Jiazheng Xu$^{1}$, Xiaohui Xie$^{1}$, Wentao Han$^{1}$, Hongning Wang$^{1,\ddagger}$}
\affiliation{%
$^{1}$ Department of Computer Science and Technology, Tsinghua University\\
$^{2}$ College of AI, Tsinghua University
}
\contribution{$^{*}$ Core Contributor; $^{\dagger}$ Project Lead; $^{\ddagger}$ Corresponding Author. }
\metadata[Contact]{\email{yks23@mails.tsinghua.edu.cn}, \email{lql24@mails.tsinghua.edu.cn}}

\newif\ifnamed
\namedtrue

\newif\ifcameraready
\ifnamed\camerareadytrue\else\camerareadyfalse\fi

\metadata[Project page]{\url{https://aaarena.net}}
\metadata[Code]{\url{https://github.com/THU-CST-SAST/AAArena}}

\abstract{Adversarial games have driven advances from heuristic search to
reinforcement learning, yet learning and adapting strategies from limited samples
remain challenging. AI agents offer an alternative by turning game experience
into revisions of executable policies. Building on heuristic learning (HL), we
formalize Adversarial Heuristic Learning (AHL), a paradigm that uses AI agents as
learning engines to refine game policies and supporting software while keeping
model weights fixed. We introduce AAArena, a benchmark comprising \val{ngames}
authentic adversarial games and 1,920 archived human programs, with an evaluation
protocol modeled on real-world game competitions. Agents interpret rules, choose
opponents, analyze replays, and revise game agents to achieve their highest ranking
within fixed match and evaluation budgets. We evaluate \val{completedmodels} model
and harness configurations: Opus5.5 with Claude Code earns
\val{bestconfigrankonegames} gold medals (rank-1 finishes), while no evaluated
configuration tops the remaining \val{unconqueredgames} human ladders. Performance
is generally weaker in games with more complex rule specifications. Further
experiments show that opponent selection and dense feedback support policy
improvement, and that agents learn from both on-policy replays of their own matches
and off-policy replays of other players' matches. These results highlight HL's
potential in adversarial games and identify persistent challenges in game
understanding, strategy implementation, and long-horizon policy development.}

\begin{document}

\maketitle

\section{Introduction}
\label{sec:introduction}

Adversarial games have long served as a testing ground for artificial intelligence. Early
systems relied on hand-crafted heuristics, evaluation functions, and tree search;
reinforcement learning and self-play subsequently advanced experience-driven policy
optimization \citep{deepblue2002,coulom2006mcts,alphazero2018}.
Opponents form part of each player's environment, and changes in their strategies
shift the learning target. Adapting and revising strategies in a changing
environment with limited interaction remains a challenge
\citep{hernandezleal2019survey}.

AI agents now work in code repositories
\citep{sweagent2024}.
They also operate graphical interfaces
\citep{osworld2024}, while agent-design and recursive program-improvement methods
let models revise the software supporting their behaviour
\citep{adas2024,stop2023}.
These capabilities suggest that AI agents can serve as learning engines.
In \emph{Learning Beyond Gradients}, \citet{weng2026learning_beyond_gradients}
describes \emph{heuristic learning} (HL), in which an AI agent accumulates
experience in editable software structures.
The agent reads failures, logs, and replays to revise programs and their supporting software.
Recent work also studies program-policy revision and tournament-based code evolution
\citep{garnier2026heuristic,evopolicygym2026,catarena2025,codeclash2025}.

The Tsinghua University Agent Competition is an annual programming competition
in which contestants build game agents to compete in adversarial games
\citep{saiblo}.
Each year, organizers introduce a new game and publish its rules.
Contestants interpret the rules, design executable strategies, and improve
their programs through freely chosen matches to compete for leaderboard positions.

\begin{samepage}
\textbf{Can AI agents undertake the same process, learning from rules and game
feedback to continually improve the game agents they build?}
This requires translating game understanding into executable policies and planning
tests and revisions within a limited budget.\par
\end{samepage}

We formalize this feedback-driven program improvement as \emph{Adversarial Heuristic
Learning} (AHL), an instantiation of HL in adversarial tasks.
The AI agent revises code policies and supporting software from game experience;
the game agent executes these policies to compete. The base model's weights remain fixed.

We build \emph{Agents for Agents Arena} (\benchname) from the competition's past
games, runnable backends, and archived human programs, turning this development
process into a unified evaluation task.
Across \val{ngames} original games, \benchname budgets selected matches and full-pool
evaluations separately to assess whether AI agents can improve their programs
and climb human-program ladders.

Our strongest configuration, Opus5.5 with Claude Code, earns \val{bestconfigrankonegames} gold medals
(rank 1 in the frozen human ladders) across \val{ngames} games; no evaluated model tops
the remaining \val{unconqueredgames} ladders.

\begin{figure}[t]
\centering
\includegraphics[width=\linewidth]{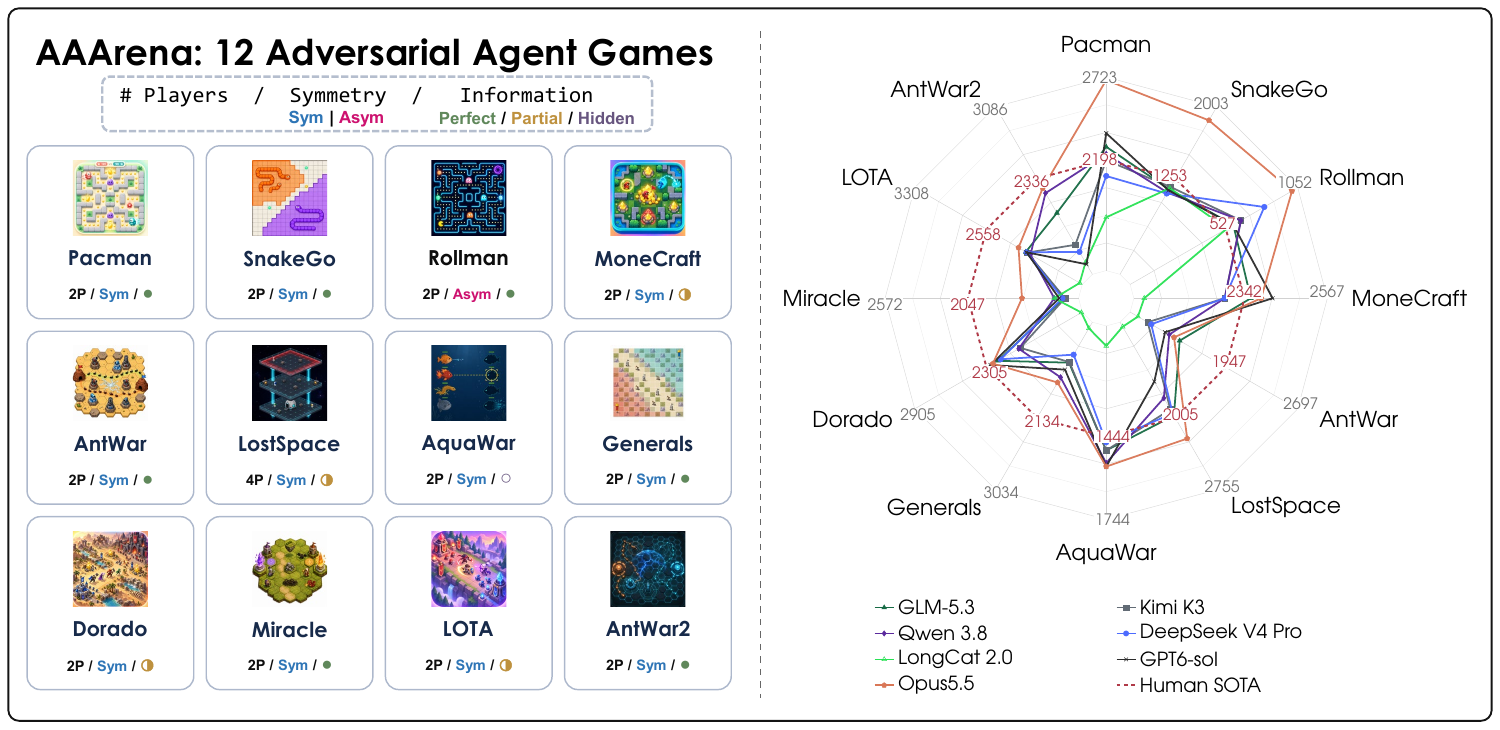}
\caption{\textbf{AAArena overview.} Left: the \val{ngames} games in ascending
rule-description size (abstract syntax tree node count), with player counts, role symmetry, and information structure
(Table~\ref{tab:games}). Right: retained-champion Elo for seven models and human SOTA.
Each spoke reports native Elo, with game-specific tick intervals; higher Elo lies
farther outward. Human SOTA forms the dashed regular dodecagon, with its Elo labeled
in red. Model colours match Figure~\ref{fig:main-bars}.}
\label{fig:hl-arena-overview}
\end{figure}

Our contributions are threefold.
\begin{itemize}[leftmargin=*]
  \item \textbf{We formalize adversarial heuristic learning (AHL)},
  a paradigm that uses AI agents as learning engines to learn from
  adversarial matches and iteratively refine executable game policies.
  \item \textbf{We introduce \benchname}, a benchmark comprising
  12 authentic adversarial games and 1,920 archived human programs,
  with an evaluation protocol modeled on real-world adversarial
  game competitions.
  \item \textbf{We evaluate seven AI model--harness configurations on
  \benchname}, with the strongest topping six human ladders, while the
  remaining six games generally have more complex rules and remain untopped
  by any evaluated configuration. Further experiments show the importance
  of opponent selection and dense feedback, and the potential of off-policy
  experience from other players' replays to overcome learning plateaus.
  Case studies reveal persistent
  challenges in game understanding, strategy implementation, and long-horizon
  policy development.
\end{itemize}

\section{Adversarial Heuristic Learning}
\label{sec:hl}

\begin{figure}[!htbp]
\centering
\includegraphics[width=\linewidth]{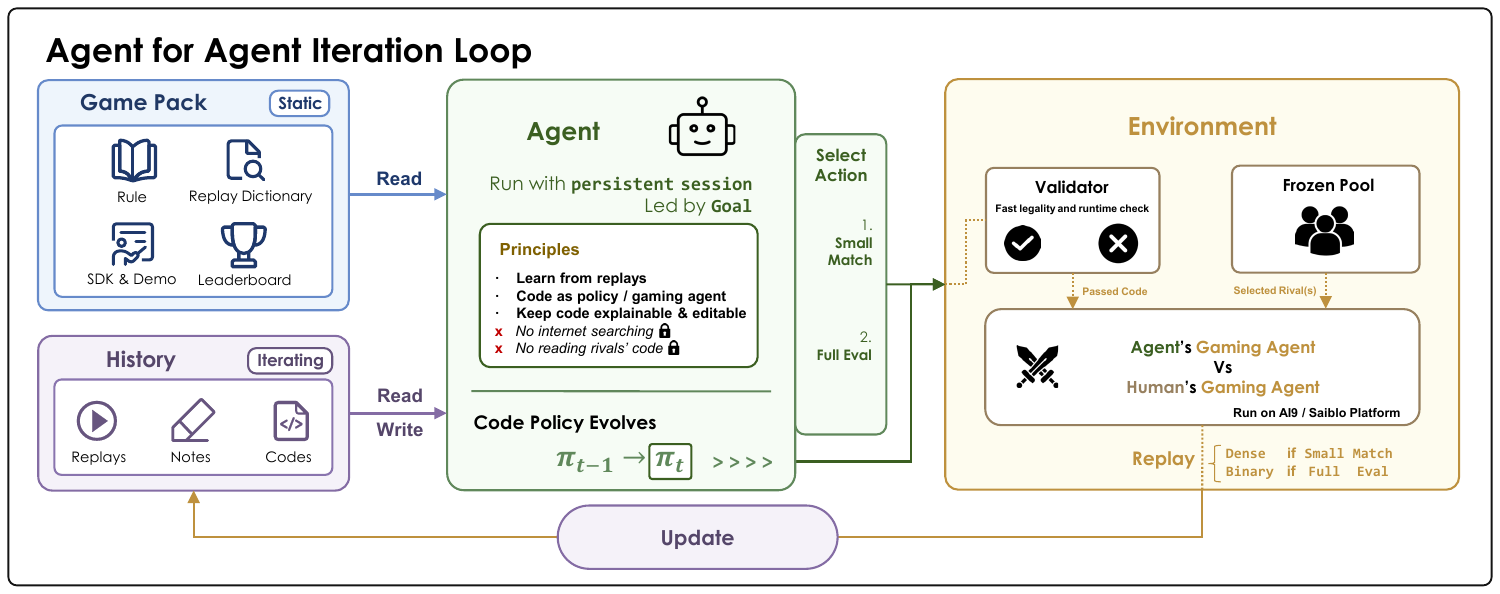}
\caption{\textbf{Agent for agent iteration loop.} An AI agent uses game resources,
replays, and retained notes to revise an executable game-agent policy. Validated
snapshots enter small matches against selected opponents or full-pool evaluations
against the frozen pool. Small matches return dense replays; full-pool evaluations
return outcomes, Elo, and rank. The agent retains development history across revisions.}
\label{fig:ahl-iteration-loop}
\end{figure}

\subsection{Preliminaries}
\label{sec:game-policies}

\paragraph{Game agent.}
An adversarial game $\mathcal{G}$ has a shared environment and $n\geq2$ game agents,
with conflicting objectives for at least two agents.
Games can assign asymmetric roles and restrict each agent's observations.
Each game agent $i$ executes a \emph{code policy} $P_i$ that selects actions as
\[
  a_t^i \sim \pi_{P_i}(\cdot\mid h_t^i),
\]
where $h_t^i$ is its permitted observation history and $a_t^i$ is its action.
The environment processes these actions to determine the outcome, while the
submitted program acts autonomously throughout the match.

\paragraph{AI agent.}
An AI agent $\mathcal{A}_{\theta}$ develops the game agent's policy between
matches; $\theta$ denotes the base model's parameters.
It reads game resources, source code, and feedback to revise the policy and its
supporting software. In \emph{Agents for Agents}, the resulting game agent's
competitive performance measures the AI agent's policy-development ability.

\paragraph{Heuristic learning.}
\emph{Heuristic learning} (HL) incorporates experience through an AI agent's
direct revisions to a software system \citep{weng2026learning_beyond_gradients}.
Feedback may include rewards, tests, logs, or replays.
The agent can revise policy logic, state representations, tests, configuration,
and memory. We call the executable policy and its supporting software and records
the \emph{heuristic system}.

\subsection{Heuristic learning in adversarial games}
\label{sec:hl-method}

\begin{definition}[Adversarial heuristic learning]
Adversarial heuristic learning (AHL) is an instantiation of HL in adversarial tasks.
An AI agent uses experience and feedback to revise a heuristic system and improve
competitive performance, while the base model's parameters $\theta$ remain fixed.
\end{definition}

The heuristic system consists of an executable policy $P$ and development
state $C$, which stores supporting tools, tests, experiment records, and memory.
Figure~\ref{fig:ahl-iteration-loop} summarizes the interaction loop;
Algorithm~\ref{alg:ahl-policy-development} gives its formal form.
The AI agent acquires game experience, interprets feedback, and revises the
heuristic system. It chooses how to collect experience and organize revisions;
the algorithm leaves feedback sources, evaluation schedules, and stopping
criteria to the task setting.

\begin{algorithm}[!htbp]
\caption{Adversarial Heuristic Learning (AHL)}
\label{alg:ahl-policy-development}
\begin{algorithmic}[1]
\Statex \textbf{Input:} AI agent $\mathcal{A}_{\theta}$ with fixed parameters $\theta$; game resources $R$;
\Statex initial policy $P_0$ and development state $C_0$
\State $(P,C)\gets(P_0,C_0)$
\While{the stopping criterion remains unmet}
  \State Acquire game experience through interactions or available records
  \State Interpret feedback using $\mathcal{A}_{\theta}$ and $(R,P,C)$
  \State Revise $P$ and its supporting state $C$ from the feedback
\EndWhile
\State \textbf{Output:} executable policy $P$
\end{algorithmic}
\end{algorithm}
\FloatBarrier

\section{\benchname}
\label{sec:benchmark}
\label{sec:arena-formulation}

\subsection{Games and opponent pools}
\label{sec:arena-provenance}
\label{sec:arena-archive}

The \val{ngames} competition games cover tactical combat, territory control,
resource management, and survival. Each has a verified, frozen opponent pool
(Table~\ref{tab:games}). Agents receive rules, SDKs, replay documentation, a leaderboard,
and a permitted reference program. Appendices~\ref{app:games} and~\ref{app:arena-protocol}
describe the games and resources.

\begin{table}[tbp]
\centering\small
\begin{tabularx}{\linewidth}{@{}lXrrr@{}}
\toprule
Game & Strategic setting & \shortstack{Pool\\programs} & \shortstack{AST\\nodes} & \shortstack{Rules\\(RA)} \\
\midrule
Pacman & Competitive maze collection & \val{pool-pacman} & 1,127 & 219 \\
SnakeGo & Snake movement and territory & \val{pool-snakego} & 1,538 & 308 \\
Rollman & Asymmetric maze pursuit & \val{pool-rollman} & 1,539 & 312 \\
MoneCraft & Mining and resource control & \val{pool-monecraft} & 1,823 & 347 \\
AntWar & Tower defence and economy & \val{pool-antwar} & 2,160 & 368 \\
LostSpace & Multiplayer survival and escape & \val{pool-lostspace} & 2,297 & 426 \\
AquaWar & Hidden-identity tactical combat & \val{pool-aquawar} & 2,531 & 439 \\
Generals & Territorial and army control & \val{pool-generals} & 2,683 & 461 \\
Dorado & Resource competition and army development & \val{pool-dorado} & 4,241 & 616 \\
Miracle & Hex-grid unit tactics & \val{pool-miracle} & 4,391 & 581 \\
LOTA & Hero control and lane combat & \val{pool-lota} & 4,513 & 684 \\
AntWar2 & Ordered tower-defence operations & \val{pool-antwar2} & 6,368 & 1,008 \\
\bottomrule
\end{tabularx}
\caption{The \val{ngames} games, ordered by increasing abstract syntax tree (AST)
node count for their rule representations. RA counts rule atoms, or independently
changeable rule propositions. Pool sizes count executable programs, not unique
human participants.}
\label{tab:games}
\end{table}

\paragraph{Rule-description size.}
Kolmogorov complexity $K(R)=\min_{p:U(p)=R}|p|$ is the length of the shortest
program describing $R$ on a fixed universal machine $U$ \citep{vitanyi2020incomputable}.
We report $C_{\mathrm{AST}}(G)=|\mathrm{AST}(R_G)|$, the node count of the rules'
abstract syntax tree (AST), and rule atoms (RA), the count of independently
changeable rule propositions. These representation-dependent counts describe
rule specifications; they do not estimate exact Kolmogorov complexity or
optimal-play difficulty. The same rule-analysis procedure gives \textit{Go}
501 AST nodes and 125 rules, and \textit{chess} 1,551 AST nodes and 238 rules.
The suite spans 1,127--6,368 AST nodes and 219--1,008 rules.

\subsection{Interaction interfaces}
\label{sec:arena-eval}

\textbf{Small matches} support targeted diagnosis by letting the AI agent select
opponents and inspect detailed replays. \textbf{Full-pool evaluations} measure
competitive progress against the complete opponent pool and return Elo, rank,
and per-opponent outcomes. The first channel concentrates feedback on specific
strategic hypotheses; the second supplies a common reference for comparing snapshots.
Both evaluate immutable programs through the same game referee and return feedback
for subsequent revisions.

\subsection{Evaluation protocol}
\label{sec:arena-protocol}

Each run receives two budgets: $B_s$ small-match units (one per requested opponent)
and $B_\ell$ full-pool evaluations. Default limits are
$B_s=\val{smallbudget}$ and $B_\ell=\val{largebudget}$. Within these limits, the agent
chooses when to revise, validate locally, or request either kind of feedback.

\textbf{The final result is the highest officially evaluated Elo attained during
the run, together with its corresponding rank.} If $\mathcal{V}$ is the set of
officially evaluated snapshots, including the common initialization baseline, then
\[
  \operatorname{Score}(B_s,B_\ell)
  =\max_{P\in\mathcal{V}}\widehat r(P).
\]
We retain the policy attaining this score, allowing later revisions to explore
without discarding a stronger snapshot. Opponent Elo ratings stay fixed throughout the run.
Appendix~\ref{app:arena-protocol} specifies ranking, budget accounting, and the
baseline's separate allowance.

\section{Experiments}
\label{sec:experiments}

\subsection{Performance of AI Agents Developing Game Agents}
\label{sec:main-results}
\label{sec:exp-human}
\begin{table}[!htb]
\centering
\caption{\textbf{Main results across 12 adversarial games, ordered by rule complexity.} Rule complexity uses AST node counts (Section~\ref{sec:arena-provenance}). A green check marks at least one configuration reaching rank 1; a red cross marks none. Each model entry gives Elo ($\uparrow$) and rank ($\downarrow$) under a 128-small/16-full budget; a gold medal marks rank 1. Bold marks the best value per row. All models use \texttt{max} reasoning effort. Opus5.5 uses Claude Code; others use Codex.}
\label{tab:main}
\label{tab:human-results}
\begingroup
\setlength{\tabcolsep}{1.3pt}
\renewcommand{\arraystretch}{1.12}
\fontsize{7.5}{9}\selectfont
\begin{tabular*}{\textwidth}{@{\extracolsep{\fill}}lrc*{7}{rr@{\hspace{8pt}}}}
\multicolumn{17}{l}{\footnotesize Rule complexity: low (top) to high (bottom) $\downarrow$} \\[3pt]
\toprule
Game & \shortstack{Rule\\complexity} & Topped & \multicolumn{2}{c}{Opus5.5} & \multicolumn{2}{c}{GPT6-sol} & \multicolumn{2}{c}{GLM-5.3} & \multicolumn{2}{c}{Kimi K3} & \multicolumn{2}{c}{\shortstack{DeepSeek\\V4 Pro}} & \multicolumn{2}{c}{Qwen 3.8} & \multicolumn{2}{c}{LongCat 2.0} \\
\cmidrule(lr){4-5}\cmidrule(lr){6-7}\cmidrule(lr){8-9}\cmidrule(lr){10-11}\cmidrule(lr){12-13}\cmidrule(lr){14-15}\cmidrule(lr){16-17}
& & & Elo & Rank & Elo & Rank & Elo & Rank & Elo & Rank & Elo & Rank & Elo & Rank & Elo & Rank \\
\midrule
Pacman & 1,127 & \textcolor{green!50!black}{$\checkmark$} & \textbf{2703.9} & \textbf{1}\goldrank & 2367.6 & 1\goldrank & 2281.6 & 1\goldrank & 2196.3 & 2 & 2096.8 & 4 & 2236.1 & 1\goldrank & 1835.0 & 10 \\
SnakeGo & 1,538 & \textcolor{green!50!black}{$\checkmark$} & \textbf{1860.9} & \textbf{1}\goldrank & 1129.6 & 4 & 1141.2 & 4 & 1165.9 & 3 & 1097.0 & 5 & 1107.5 & 5 & 1153.3 & 4 \\
Rollman & 1,539 & \textcolor{green!50!black}{$\checkmark$} & \textbf{1010.2} & \textbf{1}\goldrank & 581.2 & 1\goldrank & 581.2 & 1\goldrank & 636.9 & 1\goldrank & 807.7 & 1\goldrank & 636.9 & 1\goldrank & 534.2 & 1\goldrank \\
MoneCraft & 1,823 & \textcolor{green!50!black}{$\checkmark$} & 2387.1 & 1\goldrank & \textbf{2418.2} & \textbf{1}\goldrank & 2359.2 & 1\goldrank & 2288.2 & 3 & 2288.2 & 3 & 2288.2 & 3 & 2069.9 & 9 \\
AntWar & 2,160 & \textcolor{red!80!black}{$\times$} & 1405.1 & 5 & 1311.7 & 6 & \textbf{1464.2} & \textbf{5} & 1130.2 & 9 & 1167.2 & 9 & 1355.0 & 5 & 1028.3 & 11 \\
LostSpace & 2,297 & \textcolor{green!50!black}{$\checkmark$} & \textbf{2220.2} & \textbf{1}\goldrank & 1625.2 & 7 & 1980.1 & 2 & 1916.1 & 2 & 1956.8 & 2 & 1797.7 & 3 & 1052.9 & 77 \\
AquaWar & 2,531 & \textcolor{green!50!black}{$\checkmark$} & \textbf{1552.5} & \textbf{1}\goldrank & \textbf{1552.5} & \textbf{1}\goldrank & 1494.1 & 1\goldrank & 1494.1 & 1\goldrank & 1464.5 & 1\goldrank & 1539.7 & 1\goldrank & 1116.1 & 33 \\
Generals & 2,683 & \textcolor{red!80!black}{$\times$} & \textbf{1689.6} & \textbf{10} & 1528.4 & 13 & 1441.2 & 17 & 1433.3 & 17 & 1340.8 & 23 & 1624.9 & 10 & 1010.3 & 58 \\
Dorado & 4,241 & \textcolor{red!80!black}{$\times$} & 2252.4 & 2 & \textbf{2271.5} & \textbf{2} & 2210.3 & 2 & 2023.4 & 16 & 2187.8 & 3 & 2035.8 & 14 & 1512.4 & 156 \\
Miracle & 4,391 & \textcolor{red!80!black}{$\times$} & \textbf{1704.8} & \textbf{6} & 1469.5 & 20 & 1436.8 & 26 & 1432.8 & 26 & 1448.8 & 24 & 1500.2 & 18 & 1500.2 & 18 \\
LOTA & 4,513 & \textcolor{red!80!black}{$\times$} & \textbf{2224.2} & \textbf{15} & 2140.3 & 22 & 2147.4 & 22 & 2133.2 & 22 & 2140.3 & 22 & 2099.0 & 25 & 1585.1 & 74 \\
AntWar2 & 6,368 & \textcolor{red!80!black}{$\times$} & \textbf{2234.4} & \textbf{5} & 1440.3 & 34 & 1977.1 & 9 & 1644.2 & 19 & 1569.6 & 25 & 2183.5 & 6 & 1460.0 & 31 \\
\midrule
\multicolumn{3}{l}{\textbf{Gold medals $\uparrow$}} & \multicolumn{2}{c}{\textbf{6}} & \multicolumn{2}{c}{4} & \multicolumn{2}{c}{4} & \multicolumn{2}{c}{2} & \multicolumn{2}{c}{2} & \multicolumn{2}{c}{3} & \multicolumn{2}{c}{1} \\
\bottomrule
\end{tabular*}
\endgroup
\end{table}

\paragraph{Setup.}
We evaluate seven model--harness configurations with the default 128-small/16-full
budgets and retained-champion scoring from Section~\ref{sec:arena-protocol}.
All models use \texttt{max} reasoning effort. Opus5.5 uses Claude Code; the other
models use Codex, so the comparison includes both model and harness differences.
Runs stop upon reaching rank 1. We repeat each setting three times and report
the median result; each run contributes its best officially evaluated policy.
Main runs use \emph{on-policy} replays from matches involving the learner's own
policy.\footnote{Here, on-policy and off-policy label replay sources, not
reinforcement-learning algorithm classes.}
Appendix~\ref{app:coding-workflow} describes the implementation.
We use GLM-5.3 for subsequent analyses because its complete interaction logs
and fixed harness support comparisons across runs.

\begin{figure*}[!htb]
\centering
\includegraphics[width=\textwidth]{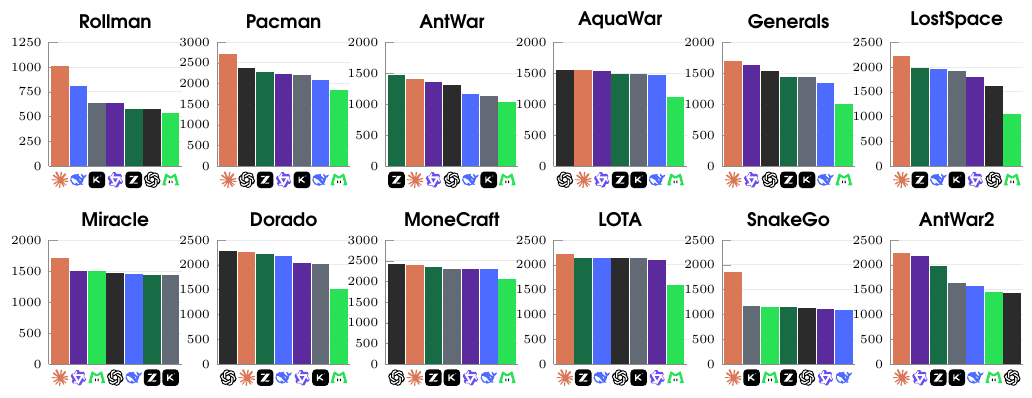}
\caption{\textbf{Per-game Elo of the retained champion.} Each game orders model configurations by Elo.}
\label{fig:main-bars}
\end{figure*}

\paragraph{Results.}
AI-developed game agents top six frozen human ladders; every evaluated
configuration falls short of rank 1 on the other six
(Table~\ref{tab:main} and Figures~\ref{fig:hl-arena-overview} and~\ref{fig:main-bars}).
Opus5.5 with Claude Code earns \val{bestconfigrankonegames} gold medals and
the highest or tied-highest Elo on nine games.
Several configurations reach rank 1 on \textit{Rollman} and \textit{AquaWar},
whereas only Opus5.5 with Claude Code does so on \textit{LostSpace} and \textit{SnakeGo}.

The remaining gaps also differ across games. The best policy ranks second on
\textit{Dorado}, while the best policies on \textit{LOTA}, \textit{Generals}, and
\textit{Miracle} remain outside the top five.
Gold medals summarize rank-1 coverage; per-game Elo and rank preserve these
differences without averaging across game-specific rating scales.

Five of the six games with smaller rule descriptions have a rank-1 finish,
compared with one of the six games with larger descriptions
(Table~\ref{tab:main}). This pattern links larger rule
specifications with gaps in competitive policy development.

\paragraph{Takeaway.}
The strongest configuration surpasses human champions in six games, while
challenging games remain beyond the evaluated agents. Performance is generally
weaker in games with more complex rule specifications, highlighting the challenge
of translating rule understanding into competitive policies.

\subsection{How Do Policies Evolve During Learning?}
\label{sec:case-study}
\label{sec:exp-long}

\subsubsection{Evolution Patterns Across Games}
We examine the 12 GLM-5.3 runs under the default 128-small/16-full budget
to distinguish how policies improve across games. Each trajectory records
the Elo and rank of individual full-pool submissions, the order of small-match
and full-pool requests, and token use. Submission scores can fall after a
revision even though the retained-champion score keeps the best result.
The trajectories follow three patterns:

\textbf{Success: reaching rank 1.} \textit{Rollman}, \textit{MoneCraft},
\textit{Pacman}, and \textit{AquaWar} reach rank 1 and stop before exhausting
the full-pool evaluation budget. These runs show that feedback-driven program
revision can produce policies that top the human-program ladders within the
given interaction limits.

\textbf{Progress: gains with regressions.} \textit{AntWar}, \textit{Dorado},
\textit{LostSpace}, \textit{SnakeGo}, and \textit{AntWar2} improve overall,
but individual revisions can reduce performance. \textit{Dorado} makes several
consecutive gains in the middle of its run, whereas \textit{AntWar2} suffers
a sharp intermediate drop and \textit{LostSpace} ends with prolonged
fluctuations. Retaining the best evaluated policy preserves earlier gains
when later edits regress.

\textbf{Stagnation: persistent plateaus.} \textit{Generals}, \textit{Miracle},
and \textit{LOTA} show extended intervals with little improvement after
initial adaptation. Further revisions and evaluations repeatedly return
to similar performance levels, showing that continued interaction does
not consistently yield stronger policies within these runs.

Appendix~\ref{app:main-trajectories} presents the complete trajectories in
Figures~\ref{fig:patterns-early}--\ref{fig:patterns-stagnant}.
We next trace the code changes behind \textit{Dorado}'s progress to show
how match feedback leads to specific policy revisions.

\subsubsection{Case Study: Dorado}
\begin{figure*}[!htbp]
\centering
\includegraphics[width=\textwidth]{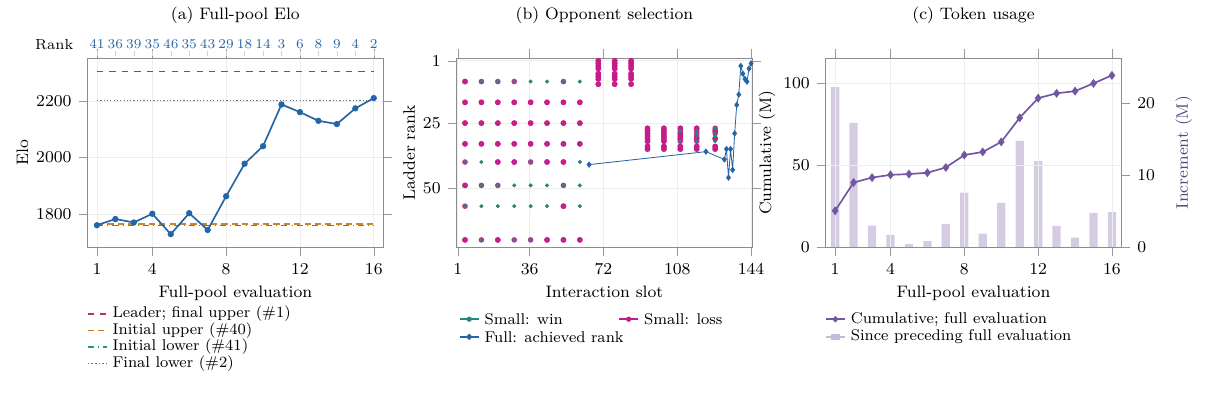}
\caption{\textbf{GLM-5.3 on \textit{Dorado}.} Full-pool Elo and rank, small--full interaction order, and token use under the 128/16 budget. Appendix~\ref{app:trajectory-plots} specifies budget-slot placement, batch markers, rank annotations, and token accounting.}
\label{fig:dorado-case}
\end{figure*}

GLM-5.3 improves its \textit{Dorado} policy from rank 41 to rank 2.
Figure~\ref{fig:dorado-case} traces its performance, and
Table~\ref{tab:dorado-milestones} shows the corresponding code changes.

\textit{Dorado} has two players, each controlling a base and recruiting heroes
from four classes. Mining supplies gold for recruitment and for upgrading
heroes near the base. Destroying the opposing base wins; simultaneous
destruction draws. If both bases remain at the turn limit, gold plus the
level-based value of each player's heroes determines the result.

\textbf{Correct mine allocation.} The central mine stays open, while other mines
have limited opening windows. Each mine uses the highest individual mining
rate; sending several heroes to one mine does not add their yields.
The agent excludes dead miners from occupancy
counts, penalizes assigning several miners to the same mine, and tracks mine
opening windows. After these revisions, the policy reaches rank 14.

\textbf{Concentrate upgrades.} The agent then upgrades two core heroes while
preserving mining income. The next full-pool evaluation raises the policy
from rank 14 to rank 3 (+147.7 Elo), the largest adjacent gain.

\textbf{Refine counterattacks.} The agent later allows a one-hero counterattack
only after sustained nearby aggression and reaches rank 2. Other attempts,
including a four-hero roster and unconditional use of the \texttt{Sacrifice}
ability, regress to ranks 46 and 43. The agent restores its best policy before
trying another revision.

\begin{table}[!htbp]
\centering
\caption{\textbf{\textit{Dorado} champion milestones.} Mechanism descriptions with added (+) and deleted (-) C++ diff excerpts. Ellipses omit context.}
\label{tab:dorado-milestones}
\footnotesize
\setlength{\tabcolsep}{4pt}
\renewcommand{\arraystretch}{1.04}
\begin{tabularx}{\textwidth}{@{}rrr>{\raggedright\arraybackslash}Xl@{}}
\toprule
\shortstack{Full-pool\\eval.} & Elo & Rank & Policy adjustment & Code excerpt \\
\midrule
2 & 1782.0 & 36 & Keep \texttt{mining}; attack only a vulnerable \texttt{enemy base}. & {\fontsize{7}{8.2}\selectfont\ttfamily\renewcommand{\arraystretch}{.90}\begin{tabular}[t]{@{}l@{}}\textcolor{green!35!black}{+232 enemyBase != NULL \&\& enemyBase->hp < 1800} \\ \textcolor{green!35!black}{     \&\& (int)myHeroes.size() >= 4;}\end{tabular}} \\[3pt]
4 & 1800.7 & 35 & Use all \texttt{living heroes} to defend a serious threat. & {\fontsize{7}{8.2}\selectfont\ttfamily\renewcommand{\arraystretch}{.90}\begin{tabular}[t]{@{}l@{}}\textcolor{green!35!black}{+245 int count = seriousThreat} \\ \textcolor{green!35!black}{     ? (int)defenders.size() : 2;}\end{tabular}} \\[3pt]
6 & 1802.8 & 35 & Trigger \texttt{full defence} for active threats or recent damage. & {\fontsize{7}{8.2}\selectfont\ttfamily\renewcommand{\arraystretch}{.90}\begin{tabular}[t]{@{}l@{}}\textcolor{green!35!black}{+228 const bool seriousThreat = ...} \\ \textcolor{green!35!black}{     || baseThreat || recentBaseHit;}\end{tabular}} \\[3pt]
8 & 1862.8 & 29 & Exclude \texttt{dead miners} from occupancy; spread assignments. & {\fontsize{7}{8.2}\selectfont\ttfamily\renewcommand{\arraystretch}{.90}\begin{tabular}[t]{@{}l@{}}\textcolor{green!35!black}{+343 if (alive \&\& mineUsable(it->second, round))} \\ \textcolor{green!35!black}{         occ[it->second]++;} \\ \textcolor{red!65!black}{-477 ... + 600LL * occ[m];} \\ \textcolor{green!35!black}{+472 ... + 2000LL * occ[m];}\end{tabular}} \\[3pt]
9 & 1977.9 & 18 & Penalize duplicate \texttt{mine assignments} more strongly. & {\fontsize{7}{8.2}\selectfont\ttfamily\renewcommand{\arraystretch}{.90}\begin{tabular}[t]{@{}l@{}}\textcolor{red!65!black}{-472 ... + 2000LL * occ[m];} \\ \textcolor{green!35!black}{+472 ... + 5000LL * occ[m];}\end{tabular}} \\[3pt]
10 & 2040.1 & 14 & Track \texttt{mine opening windows} and prioritize \texttt{open mines}. & {\fontsize{7}{8.2}\selectfont\ttfamily\renewcommand{\arraystretch}{.90}\begin{tabular}[t]{@{}l@{}}\textcolor{green!35!black}{+217 s\_mineExistUntil[i] =} \\ \textcolor{green!35!black}{         round + exist->val[0] - 1;} \\ \textcolor{green!35!black}{+ 36 \#define MINE\_OPEN\_OCCUPANCY\_PENALTY 50000LL}\end{tabular}} \\[3pt]
11 & 2187.8 & 3 & Concentrate \texttt{upgrades} on a \texttt{two-hero core}; preserve \texttt{mining income}. & {\fontsize{7}{8.2}\selectfont\ttfamily\renewcommand{\arraystretch}{.90}\begin{tabular}[t]{@{}l@{}}\textcolor{green!35!black}{+296 if (s\_dutyId < 0} \\ \textcolor{green!35!black}{         \&\& (int)s\_leveledCore.size() < 2 ...)} \\ \textcolor{green!35!black}{+316 if (!coreHammerguard \&\&} \\ \textcolor{green!35!black}{         type == "hammerguard") priority = 0;}\end{tabular}} \\[3pt]
16 & 2210.3 & 2 & Allow \texttt{one-hero counterattacks} only after sustained aggression. & {\fontsize{7}{8.2}\selectfont\ttfamily\renewcommand{\arraystretch}{.90}\begin{tabular}[t]{@{}l@{}}\textcolor{green!35!black}{+410 s\_aggressionRounds = tightAggression} \\ \textcolor{green!35!black}{         ? s\_aggressionRounds + 1 : 0;} \\ \textcolor{green!35!black}{+412 const bool corePush =} \\ \textcolor{green!35!black}{         activeCoreHeroes >= 2 \&\& aggressionThreat;}\end{tabular}} \\[3pt]
\bottomrule
\end{tabularx}
\end{table}

\FloatBarrier

\subsection{Does More Interaction Overcome Performance Plateaus?}
\label{sec:convergence}

We test whether more interaction resolves GLM-5.3's weak results on
\textit{Generals}, \textit{Miracle}, \textit{LOTA}, and \textit{AntWar2}.
Each continuation keeps the existing workspace and evaluation ledger and adds
256 small-match units and 32 full-pool evaluations. The total budget is 384/48,
three times the original 128/16. Table~\ref{tab:continuation} compares the best
policies at both budget levels.

\begin{table}[!htbp]
\centering
\caption{\textbf{Threefold-budget continuation for GLM-5.3.} Elo and rank of the retained champion at each budget. The extension adds 256 small units and 32 full-pool evaluations to the same run. A positive $\Delta$Rank indicates an improved position.}
\label{tab:continuation}
\small
\setlength{\tabcolsep}{7pt}
\renewcommand{\arraystretch}{1.10}
\begin{tabular*}{\textwidth}{@{\extracolsep{\fill}}lrrrrrr@{}}
\toprule
& \multicolumn{2}{c}{128 / 16} & \multicolumn{2}{c}{384 / 48} & \multicolumn{2}{c}{Change} \\
\cmidrule(lr){2-3}\cmidrule(lr){4-5}\cmidrule(lr){6-7}
Game & Elo & Rank & Elo & Rank & $\Delta$Elo & $\Delta$Rank \\
\midrule
Generals & 1441.2 & 17 & 1441.2 & 17 & +0.0 & +0  \\
Miracle & 1436.8 & 26 & 1486.8 & 19 & +50.0 & +7  \\
LOTA & 2147.4 & 22 & 2147.4 & 22 & +0.0 & +0  \\
AntWar2 & 1977.1 & 9 & 2117.0 & 6 & +139.8 & +3  \\
\bottomrule
\end{tabular*}
\end{table}

\begin{figure}[!htbp]
\centering
\begin{figgroup}{grpcontinuation}{Threefold budget}
\includegraphics[width=\linewidth]{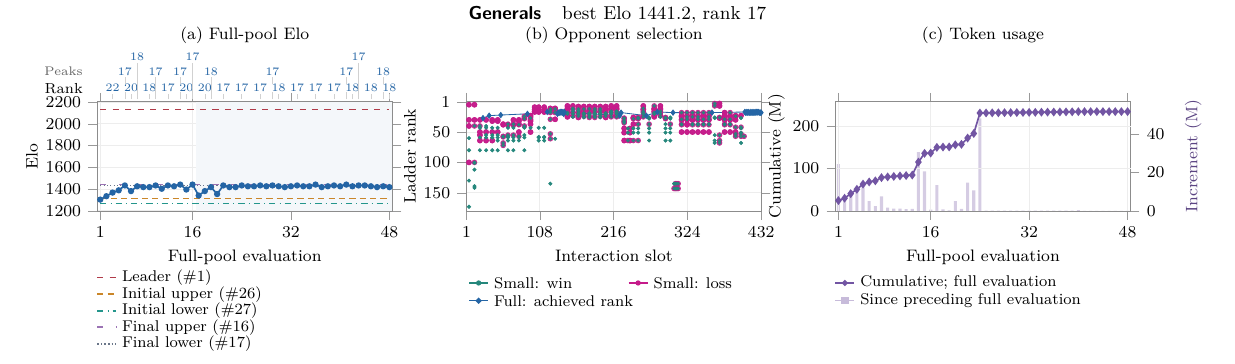}\par\smallskip
\includegraphics[width=\linewidth]{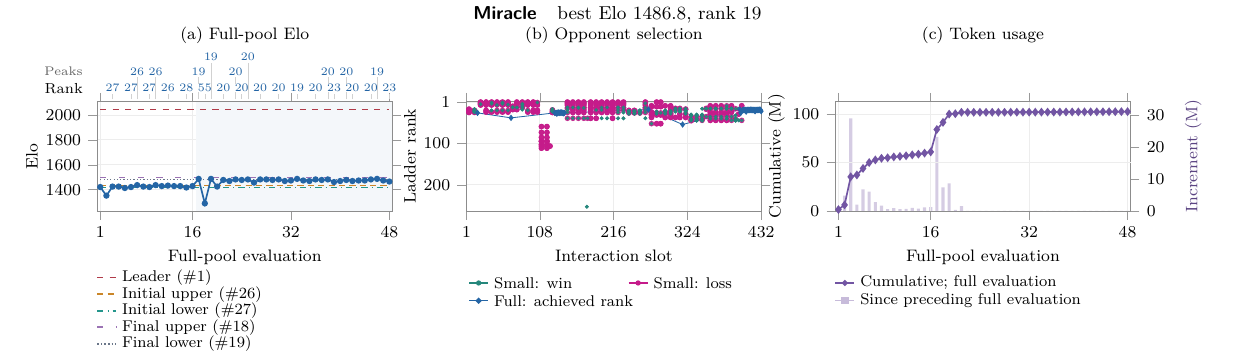}\par\smallskip
\includegraphics[width=\linewidth]{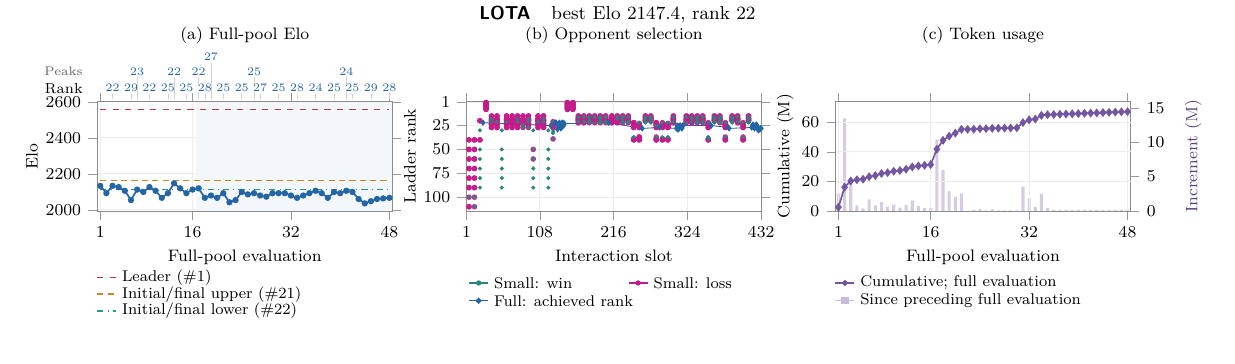}\par\smallskip
\includegraphics[width=\linewidth]{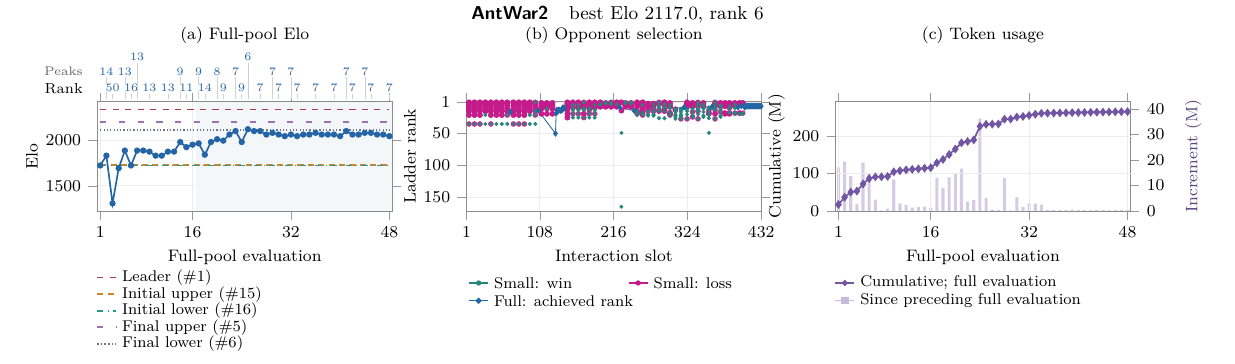}\par\smallskip
\end{figgroup}
\caption{\textbf{Policy performance under a threefold budget.} Full-pool Elo, interaction order, and token use for four continued runs. Ranks appear at every third full-pool evaluation and local Elo peaks. Plot conventions appear in Appendix~\ref{app:trajectory-plots}.}
\label{fig:continuation-trajectories}
\end{figure}

Tripling the budget leaves two retained champions unchanged and improves two;
none reaches rank 1 (Table~\ref{tab:continuation}).
\textit{Generals} and \textit{LOTA} keep their earlier best scores, while
\textit{Miracle} moves from rank 26 to 19 and \textit{AntWar2} from rank 9 to 6.
Figure~\ref{fig:continuation-trajectories} shows flat intervals and fluctuations
through the extended runs.
These outcomes describe a finite continuation of four GLM-5.3 runs under
the same development protocol.

\paragraph{Takeaway.}
Tripling the interaction budget yields limited further improvement, suggesting
that the default budget brings these runs close to a performance plateau.
Within the tested range, more interaction alone does not overcome the remaining
performance gaps.

\subsection{Which Feedback Choices Affect Policy Improvement?}
\label{sec:ablations}
We select one representative game from each trajectory category: \textit{Pacman} (Success), \textit{AntWar} (Progress), and \textit{Miracle} (Stagnation). All ablations use GLM-5.3 with max reasoning effort and the same 128-small/16-full budget. Each axis varies one component while keeping its remaining settings fixed. We report retained-champion Elo, with exact Elo/rank tables in Appendix~\ref{app:ablation-details}.

\subsubsection{Does Opponent Selection Affect Performance?}
\label{sec:opponent-ablation}
\label{sec:exp-opponent-ablation}
We compare four opponent-selection strategies. \emph{Model selection}
lets the agent choose opponents, as in the main experiment. The \emph{ladder} starts
at rank 30 and moves toward stronger opponents without returning to weaker ones.
\emph{Random sampling} draws uniformly from the pool; \emph{top-four} uses only
the four highest-ranked opponents. All conditions use dense feedback and
a small-match batch size of four.

\begin{figure}[!htbp]
\centering
\includegraphics[width=\textwidth]{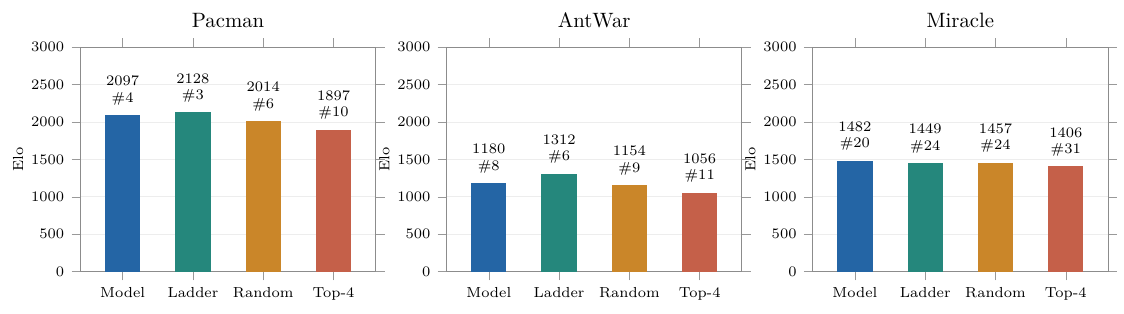}
\caption{\textbf{Opponent selection.} GLM-5.3 retained-champion Elo and rank with model, ladder, random, and top-four selection; all use dense feedback and batch size 4. Plot conventions appear in Appendix~\ref{app:trajectory-plots}.}
\label{fig:opponent-ablation}
\end{figure}

The ladder leads on \textit{Pacman} and \textit{AntWar}, while model selection
leads on \textit{Miracle} (Figure~\ref{fig:opponent-ablation}).
Model selection outperforms random and top-four selection in all three games;
the ladder trails random selection on \textit{Miracle}.
Top-four selection yields the lowest Elo throughout.
One hypothesis is that gradually stronger or agent-selected opponents
provide more actionable feedback than repeated mismatches.

\paragraph{Takeaway.}
Opponent selection is important for effective policy development: agents need
opponents whose matches provide useful information for improvement.
Agent-directed selection and progressively stronger opponents both perform well,
showing the value of choosing opponents for learning.

\subsubsection{Does Dense Replay Feedback Affect Performance?}
\label{sec:replay-ablation}
\label{sec:exp-history-ablation}
We compare detailed small-match replays and their translations with binary
win/non-win feedback. Both conditions keep conversation history, agent-selected
opponents, and full-pool feedback; the agent chooses the batch size.

\begin{figure}[!htbp]
\centering
\includegraphics[width=\textwidth]{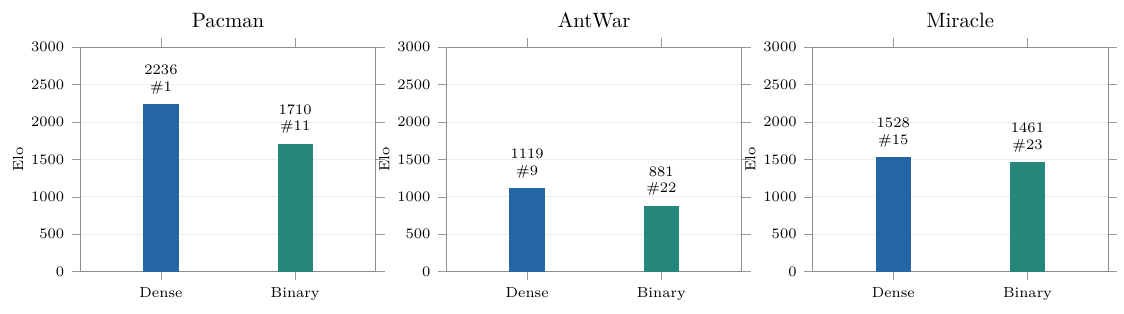}
\caption{\textbf{Replay feedback.} Dense replays versus binary win/non-win feedback under the same 128/16 budget. Labels give Elo and rank. Dense feedback improves performance in all three representative games.}
\label{fig:replay-ablation}
\end{figure}

Dense feedback yields higher retained-champion Elo in all three games,
with gains that vary by game (Figure~\ref{fig:replay-ablation}).
Detailed replays supply state, action, and consequence information for diagnosing
decision errors; binary feedback supplies the outcome alone.
The gains support retaining this diagnostic context in AHL under the tested budgets.

\paragraph{Takeaway.}
A wider feedback channel enables more efficient policy development.
Dense replays provide decision-level information that agents can use to diagnose
failures and guide code revisions, improving performance under a fixed budget.

\subsubsection{Does the Number of Opponents per Small Match Matter?}
\label{sec:opponent-count-ablation}
We compare 1, 2, 4, and 8 opponents per small-match submission, keeping model
selection and dense feedback. The total allowance remains 128 opponent tests
and 16 full-pool evaluations.

\begin{figure}[!htbp]
\centering
\includegraphics[width=\textwidth]{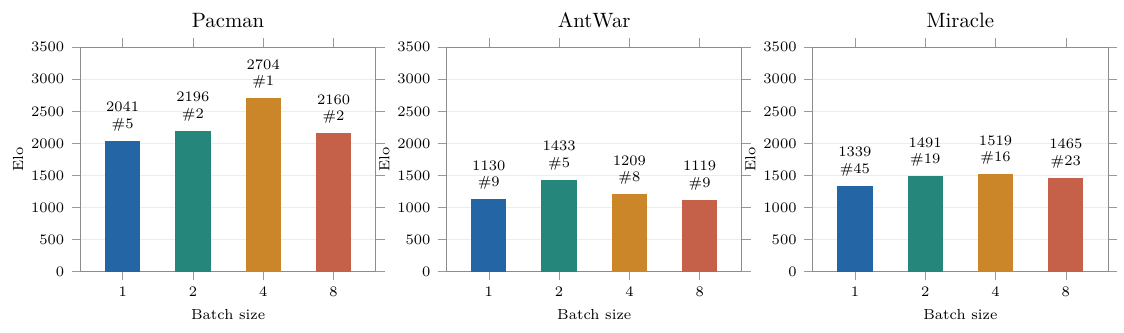}
\caption{\textbf{Small-match batch size.} Number of opponents per submission, with a fixed total allowance of 128 opponent tests and 16 full-pool evaluations. Labels give Elo and rank. The best size is 4, 2, and 4 for \textit{Pacman}, \textit{AntWar}, and \textit{Miracle}, respectively.}
\label{fig:batch-ablation}
\end{figure}

Intermediate batches perform best: four opponents on \textit{Pacman} and
\textit{Miracle}, and two on \textit{AntWar} (Figure~\ref{fig:batch-ablation}).
Eight opponents per request spend the small-match allowance in 16 full-sized
requests, compared with 32--64 requests for batches of 2--4.
The ranking is consistent with a trade-off between matchup breadth per request
and opportunities to test revised policies.
The observed preference for 2--4 opponents concerns these three games and settings.

\paragraph{Takeaway.}
The ablations show that opponent selection, replay detail, and batch size all
affect retained-champion performance under a fixed budget.
We next vary which players generate the replay evidence.

\subsection{Can Agents Learn from Off-Policy Replays?}
\label{sec:offpolicy-cloning}
\label{sec:exp-clone}
We test whether \emph{off-policy} replays from matches between other policies
support policy improvement.
On \textit{Pacman}, \textit{AntWar}, and \textit{Miracle}, GLM-5.3 chooses
replays from a catalog of matches between frozen-pool opponents.
It can request 128 replay views and 16 full-pool evaluations of its own policy.
Reasoning effort, game resources, opponent ladders, and budget ceilings match
the on-policy setting. Only the replay source changes; the agent still receives
full-pool feedback on its own policy.

\begin{figure}[!htbp]
\centering
\begin{figgroup}{grpoffpolicy}{On-policy and off-policy learning}
\includegraphics[width=\linewidth]{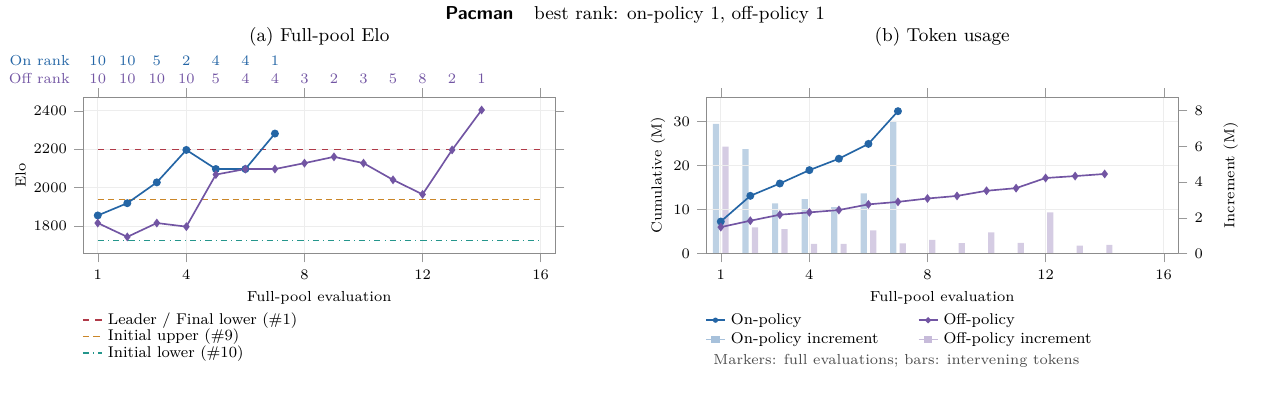}\par\smallskip
\includegraphics[width=\linewidth]{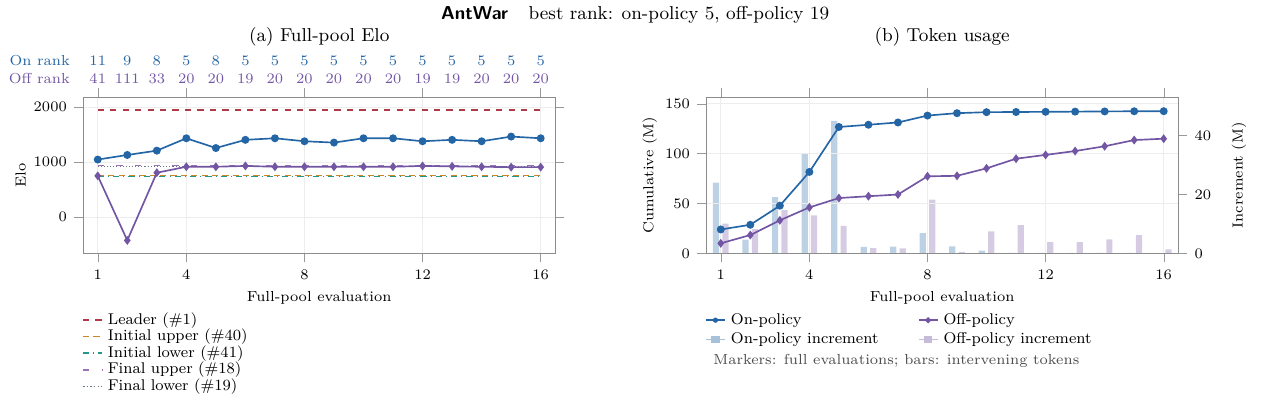}\par\smallskip
\includegraphics[width=\linewidth]{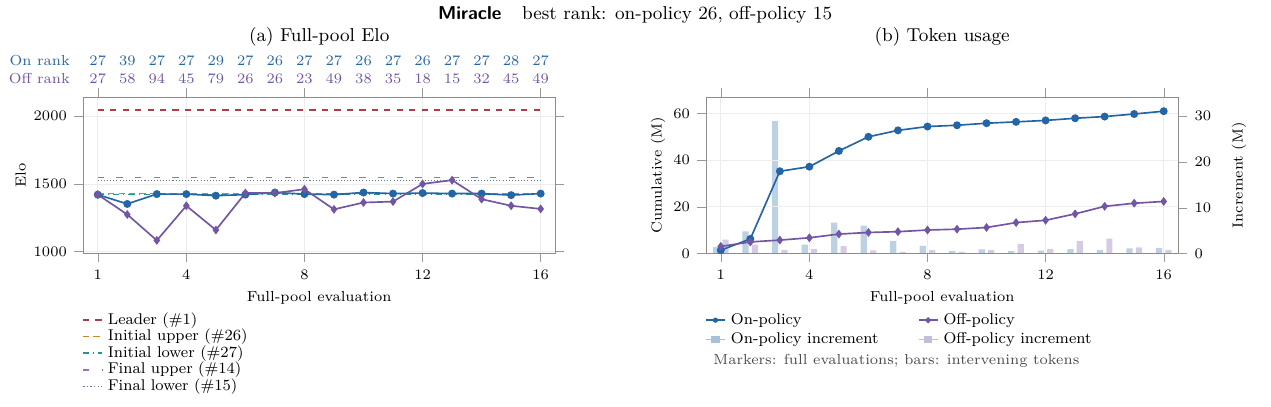}\par\smallskip
\end{figgroup}
\caption{\textbf{On-policy and off-policy learning trajectories.} Full-pool Elo and ranks, cumulative tokens, and token increments for GLM-5.3 with self-generated and other players' replays. Both conditions receive full-pool feedback on their own policies. Appendix~\ref{app:trajectory-plots} specifies rank rows, reference levels, and token accounting.}
\label{fig:offpolicy-trajectories}
\end{figure}

All three off-policy runs improve from their first evaluated policy to the
retained champion: \textit{Pacman} advances from rank 10 to 1,
\textit{AntWar} from 41 to 19, and \textit{Miracle} from 27 to 15
(Figure~\ref{fig:offpolicy-trajectories}).

\begin{figure}[!htbp]
\centering
\includegraphics[width=\textwidth]{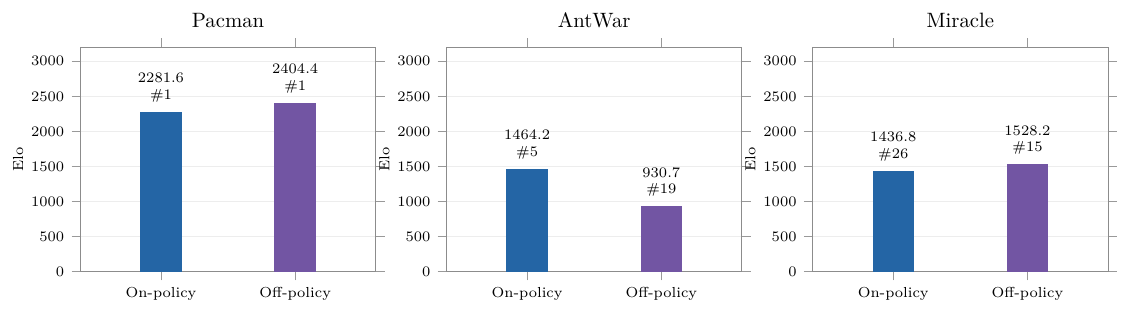}
\caption{\textbf{On-policy versus off-policy replay learning.} Retained-champion Elo and rank for GLM-5.3 under the same 128/16 budget ceilings. On-policy scores are the main-table results. Labels give Elo and rank; each panel uses the same zero-based Elo scale. Both \textit{Pacman} runs terminate early upon reaching rank 1.}
\label{fig:offpolicy-comparison}
\end{figure}

\textit{Miracle} provides the clearest contrast with learning from the agent's
own matches. Its on-policy trajectory plateaus with a retained champion at
rank 26, while the independent off-policy run reaches rank 15
(Figure~\ref{fig:offpolicy-comparison}). This contrast suggests that other
players' experience can support progress beyond the plateau observed with
self-generated replays. On \textit{Pacman}, both settings reach rank 1;
on \textit{AntWar}, on-policy learning performs better, reaching rank 5
compared with 19 off-policy.

\FloatBarrier

\paragraph{Takeaway.}
Off-policy replays broaden the experience available to AHL and can support
progress beyond the plateaus observed with on-policy learning.
Other players' experience can suggest alternative strategies, with the potential
to outperform learning from the agent's own matches.

\section{Related Work}
\label{sec:related-work}

\begin{table}[!htbp]
\centering
\caption{\textbf{Representative game-agent evaluation protocols.}
Feedback denotes information available to the evaluated AI agent:
$\checkmark$ indicates availability and a dash indicates absence in the named setting.
Traces include in-game observations or post-match logs.
CodeClash refers to its standard, closed-code tournament; GACL to code generation.}
\label{tab:related-work}
\begingroup
\small
\setlength{\tabcolsep}{3pt}
\renewcommand{\arraystretch}{1.15}
\begin{tabularx}{\linewidth}{@{}>{\raggedright\arraybackslash}X
  >{\raggedright\arraybackslash}p{.11\linewidth}
  >{\raggedright\arraybackslash}p{.20\linewidth}
  *{2}{>{\centering\arraybackslash}p{.06\linewidth}}
  >{\raggedright\arraybackslash}p{.19\linewidth}@{}}
\toprule
Work & AI output & Opponents & \multicolumn{2}{c}{Feedback} & Protocol \\
\cmidrule(lr){4-5}
 & & & \scriptsize Scores & \scriptsize Traces & \\
\midrule
\multicolumn{6}{@{}l}{\textbf{Direct play}} \\
GameBench \citep{gamebench2024}
  & Actions & Human + AI players
  & $\checkmark$ & $\checkmark$ & Direct play \\
TextArena (competitive) \citep{textarena2025}
  & Actions & Human + AI players
  & $\checkmark$ & $\checkmark$ & Direct play \\
\midrule
\multicolumn{6}{@{}l}{\textbf{Program development}} \\
GACL \citep{gacl2026}
  & Programs & AI-developed programs
  & -- & -- & One-shot \\
Vibe-coding (IR) \citep{vibecoding2025tournament}
  & Programs & Human-written + AI-developed programs
  & $\checkmark$ & -- & Preset \\
CATArena (games) \citep{catarena2025}
  & Programs & AI-developed programs
  & $\checkmark$ & $\checkmark$ & Preset \\
CodeClash (tournament) \citep{codeclash2025}
  & Programs & AI-developed programs
  & $\checkmark$ & $\checkmark$ & Preset \\
CC:Ladder \citep{codeclash2025}
  & Programs & Human-written programs
  & $\checkmark$ & $\checkmark$ & Preset \\
EvoPolicyGym \citep{evopolicygym2026}
  & Programs & None
  & $\checkmark$ & $\checkmark$ & Agent-scheduled \\
\midrule
\textbf{AAArena (ours)}
  & Programs & Human-written programs
  & $\checkmark$ & $\checkmark$ & Agent-scheduled \\
\bottomrule
\end{tabularx}
\par\smallskip
\begin{minipage}{\linewidth}
\footnotesize
Programs are executable code policies. Preset means fixed refinement rounds,
tournament rounds, or opponent order; Agent-scheduled means that the agent chooses
evaluation requests within a budget. These labels describe the outer schedule.
Scores include outcomes, returns, and rankings; traces include observations,
match logs, and replays. CC:Ladder starts from a human solution.
IR denotes iterative refinement. Rows describe specific
evaluation conditions: GACL covers code generation, and CodeClash covers its
default code-opaque tournaments. EvoPolicyGym budgets single-agent episodes and
hides validation feedback; \benchname uses frozen human programs and returns
full-pool scores during development under a separate evaluation budget.
\end{minipage}
\endgroup
\end{table}

\paragraph{Competitive program development.}
Competitive program-development benchmarks differ in their outputs, opponents,
feedback, and revision protocols (Table~\ref{tab:related-work}).
GACL evaluates generated game agents and ProxyWar includes code repair
\citep{gacl2026,proxywar2026}; the vibe-coding tournament compares student and
model programs with a self-play refinement condition \citep{vibecoding2025tournament}.
CATArena supplies tournament logs and peer code across revision rounds
\citep{catarena2025}, while CodeClash studies competitive code evolution and
CC:Ladder advances through ranked human programs after each win \citep{codeclash2025}.
EvoPolicyGym budgets episodes in single-agent environments and uses hidden
validation \citep{evopolicygym2026}.
\benchname combines \val{ngames} custom competition games, archived human-written programs,
and agent-selected testing, with separate budgets for small matches and
full-pool evaluations.

\paragraph{HL and program evolution.}
Weng's HL incorporates experience through revisions to program policies and supporting software
\citep{weng2026learning_beyond_gradients}.
Programmatic reinforcement learning and LLM-guided program search also use programs
as policy representations \citep{pirl2018,viper2018,codetoplay2024,kuang2025generative}.
Related systems improve controllers, algorithms, and agent software
\citep{garnier2026heuristic,funsearch2024,alphaevolve2025,adas2024,stop2023},
or revise rewards and training procedures \citep{eureka2023,ai4aibench2026}.
Trajectory distillation extracts reusable skills from replay
\citep{browserbc2026}.
AHL applies HL to competitive game-agent development, where the AI agent revises
policies and supporting software with a fixed base model.

\paragraph{Direct game-playing evaluation.}
Direct game-playing benchmarks measure the model's decisions during matches.
BALROG and LMGame-Bench evaluate action selection in interactive games
\citep{balrog2024,lmgamebench2025}.
GameBench, GameArena, TextArena, GAMEBoT, ZeroSumEval, and Kaggle Game Arena extend
this evaluation to competitive players
\citep{gamebench2024,gamearena2024,textarena2025,gamebot2025,zerosumeval2025,kagglegamearena2026paper}.
These settings measure the player, whereas \benchname measures the AI agent
developing the autonomous program that plays the game.

\section{Future Work}
\label{sec:future-work}

\paragraph{Extending \benchname.}
We plan to add games, controlled rule variants, and changing opponent populations
to study adaptation to changes in the task and competition.
These extensions would broaden evaluation beyond per-game rankings to adaptation
speed, cross-game transfer, and continual learning.

\paragraph{Heuristic and reinforcement learning.}
We aim to compare how code revisions and parameter updates learn from experience
and assign credit for improvements. Such comparisons should account for feedback
access and interaction budgets. Hybrid learners could use HL to revise policy
structure and reinforcement learning to train its components.

\paragraph{From replay use to active learning.}
External replays could help agents form strategic hypotheses before choosing
opponents to test them. We plan to study how agents allocate replay views and
matches to resolve uncertainty. Agents could turn validated strategies into
reusable policy modules.

\section{Conclusion}

We present \benchname to evaluate AI agents that develop executable game policies
through adversarial heuristic learning (AHL). The benchmark combines
\val{ngames} real-world competition games, archived human programs, and a unified
two-budget protocol. Across \val{completedmodels} AI models, Opus5.5 with
Claude Code earns \val{bestconfigrankonegames} gold medals; no evaluated configuration
tops the remaining \val{unconqueredgames} human ladders.
Dense feedback supports policy improvement in the ablations, and agents also
improve by learning from both on-policy and off-policy replays.
Qualitative analysis of code revisions and trajectories points to challenges in
game understanding, strategy implementation, and long-horizon planning.
By making this development process measurable, \benchname provides a foundation
for research on heuristic learning and self-evolving AI agents.

\begin{samepage}
\section*{Acknowledgement}
We thank the many members of the Student Association of the Department of Computer Science and Technology at Tsinghua University for developing and maintaining the Tsinghua University Agent Competition over the years and for supporting this work.
We thank Z.ai for sponsoring this work. We also thank all participants in past editions of the Tsinghua University Agent Competition.
\par
\end{samepage}

\appendix

\section{The evaluated games}
\label{app:games}
\label{app:arena}
\label{app:apparatus}

The suite shares an evaluation interface, but its policies solve different decision
problems. The descriptions below refer to the public rule and SDK bundles used in the
experiments; Table~\ref{tab:games} lists pool sizes.

\paragraph{\textit{Pacman}.}
\textit{Pacman} is a separate, symmetric competitive maze game. Players collect ordinary and
\texttt{high-value beans}, avoid \texttt{monsters}, and place \texttt{mines} that affect movement choices.
Collection, collision avoidance, and disruption of the opponent jointly determine the
score. Its rules and opponent pool are distinct from \textit{Rollman}'s asymmetric contest.

\paragraph{\textit{SnakeGo}.}
\textit{SnakeGo} is a territory game in which \texttt{snakes} move, collect items, claim \texttt{cells}, and can
split or use special actions. Local collision avoidance interacts with longer-horizon
territory growth, making a safe immediate move potentially costly at settlement.

\paragraph{\textit{Rollman}.}
\textit{Rollman} is an asymmetric maze game. One side collects \texttt{beans} and uses temporary skills;
the other coordinates \texttt{ghosts} to intercept it. The roles differ in both action space
and objective, making movement planning and coordinated pursuit distinct policy problems.

\paragraph{\textit{MoneCraft}.}
\textit{MoneCraft} is a mining and resource-control game on a symmetric map. Players capture
\texttt{mines} that generate \texttt{currency} and can teleport, place \texttt{mines}, or purchase protection.
The policy trades movement and disruption against sustained resource production;
end-game \texttt{currency} and \texttt{mine ownership} determine the result under the referee's rules.

\paragraph{\textit{AntWar}.}
\textit{AntWar} combines tower defence with economic management. Players build and upgrade
\texttt{towers}, improve their \texttt{bases}, and time \texttt{super-weapons}, while \texttt{ants} follow environment-driven
paths. The policy must balance present defence with spending that changes future pressure.

\paragraph{\textit{LostSpace}.}
\textit{LostSpace} is a multiplayer survival-and-escape game on a layered map that contracts
over time. Players collect \texttt{supplies} and \texttt{keys}, use \texttt{tools}, fight or evade rivals, and
seek escape. Ranked outcomes create trade-offs between immediate combat, survival,
and the timing of an escape attempt.

\paragraph{\textit{AquaWar}.}
\textit{AquaWar} is a hidden-identity tactical game. Players select \texttt{fish}, make assertions about
hidden opponent identities, and choose attacks or type-specific skills. Roster design,
inference, and information-sensitive timing all affect the outcome.

\paragraph{\textit{Generals}.}
\textit{Generals} combines territorial expansion, army movement, neutral assets, production,
technology, and tactical abilities. An effective policy coordinates economic development
with the timing and direction of attacks rather than optimizing either in isolation.

\paragraph{\textit{Dorado}.}
\textit{Dorado} combines resource competition with \texttt{hero} and army development. Players choose
\texttt{heroes}, produce and command \texttt{units}, mine neutral resources, and invest in \texttt{upgrades}.
The policy coordinates movement, combat, harvesting, and abilities to threaten opposing
\texttt{bases} while maintaining the resources needed to continue production.

\paragraph{\textit{Miracle}.}
\textit{Miracle} is a hex-grid strategy game with heterogeneous \texttt{units} and an initial drafting
stage. A player selects \texttt{creatures} and an \texttt{artifact}, then summons, moves, attacks, and
uses abilities to destroy the opponent's \texttt{miracle}. Resource limits and unit interactions
make tactical coordination central to the policy.

\paragraph{\textit{LOTA}.}
\textit{LOTA} is a lane-based combat game. Each player chooses a \texttt{hero} and controls movement,
attacks, abilities, skill development, and \texttt{revival} while \texttt{units} advance along fixed
\texttt{lanes}. Victory depends on destroying the opposing \texttt{base} or the referee's settlement
criteria at the turn limit. Effective control couples immediate fights with sustained
pressure on defensive structures.

\paragraph{\textit{AntWar2}.}
\textit{AntWar2} extends tower-defence strategy with \texttt{ordered operation lists} whose accepted
prefix changes subsequent decisions. The agent must coordinate \texttt{tower placement},
economic development, and weapon timing while respecting sequence-dependent legality.

\section{Arena resources and evaluation details}
\label{app:arena-protocol}

\paragraph{Public resources and opponents.}
We freeze the versioned opponent programs, their ratings, and the game resources
across runs. Some entrants submitted several related versions, so pool sizes
count programs, not independent participants. The workspace includes rules,
SDKs, replay documentation, a leaderboard, and one reference program
(the rank-\val{referenceRank} entry). The agent cannot access the other opponents'
source code or evaluator internals.

\paragraph{Budget accounting.}
A small-match request tests up to \val{batchmax} opponents and costs one unit
per opponent. A full-pool evaluation costs one large unit.
These budgets constrain feedback access; source edits, model calls, tokens,
and wall-clock time have no separate limits. The starter policy's initial
evaluation is free and falls outside the charged full-pool sequence.

\subsection{AI agent workflow}
\label{app:coding-workflow}

Each model--game run uses one persistent session with a writable workspace.
Opus5.5 runs in Claude Code; the other models run in Codex.
The agent's goal is to improve competitive performance. It chooses when to read
rules, edit code, test selected opponents, and submit a full-pool evaluation.
Small matches return replays; full-pool evaluations return Elo, rank, and
per-opponent outcomes. Context compaction preserves the session but shortens
its conversation history, so the agent keeps notes to retain experience.
We report the run's highest officially evaluated Elo, including the common
initialization baseline, and its corresponding rank. The controller retains
the policy snapshot attaining that score.

\subsection{Elo and rank}
\label{sec:arena-scores}

Each full-pool evaluation estimates the Elo $\widehat r(P)$ of policy snapshot $P$
from its match outcomes using a fixed-pool Bradley--Terry model
\citep{bradleyterry1952}. Let $H$ denote the frozen opponent pool and $r_j$ the
fixed rating of opponent $j$. The candidate's score in match $m$ against
opponent $j(m)$ is $s_m\in\{0,\tfrac12,1\}$ (loss, draw, win).
The expected score at rating $r$ is
\[
 p(r,r_j)=\frac{1}{1+\val{elobase}^{(r_j-r)/\val{eloscale}}},
\]
and the estimate $\hat r$ solves
\[
 \sum_m p(\hat r,r_{j(m)})+p(\hat r,\bar r_H)
 =\sum_m s_m+\tfrac12,
 \qquad
 \bar r_H=\frac{1}{|H|}\sum_{j\in H}r_j.
\]
The extra term adds one virtual draw against the complete pool's mean rating,
keeping $\hat r$ finite for perfect win or loss records.
Opponent ratings remain fixed across snapshots and runs.
The candidate's rank is $1+|\{j\in H:r_j>\hat r\}|$.
We report each game's native Elo and rank and summarize rank-1 coverage with gold medals.

\subsection{Gold medals}
\label{app:summary-metrics}

Gold medals count rank-1 finishes against the frozen human ladders. For configuration
$m$ and game $g$, let $q_{m,g}$ denote its reported rank.
Across $G=\val{ngames}$ games, we compute
\[
 \operatorname{Gold}(m)=\sum_{g=1}^{G}\mathbf{1}\{q_{m,g}=1\}.
\]
A gold medal denotes first place under this rating protocol; it does not establish a direct head-to-head
win against the human leader or optimal play.

The pool ratings and rank-1 thresholds remain fixed when we evaluate additional
models. Gold medals summarize performance against the archived pools.
Full-pool evaluations provide feedback during optimization, so the reported results
do not measure generalization to unseen opponents. The committed summary records
pin the rating inputs and calculations; they do not independently validate the
original human-rating calibration.

\subsection{Rule complexity: AST and RA}
\label{app:rule-complexity}

We compare rule-description sizes by rewriting each game's rules in a shared
formal pseudocode language. Its domain-neutral vocabulary includes top-level
declarations (\texttt{constant}, \texttt{enum}, \texttt{entity}, \texttt{action},
\texttt{observation}, \texttt{setup}, \texttt{rule}, \texttt{terminal}), fifteen
statement types, and expressions for arithmetic, comparisons, and function calls.
Statement types include \texttt{let}, \texttt{require}, \texttt{when}, \texttt{for},
\texttt{update}, \texttt{create}, \texttt{delete}, \texttt{reveal}, and \texttt{return}.
Set membership, distance, and probability are primitives.
We explicitly define every game-specific operation, such as splitting a snake
or casting a fish skill.
Each description covers players, information, state, initialization, randomness,
legal actions, transitions, and termination. We exclude communication protocols,
rendering, logging, and platform-only timeouts.

\paragraph{AST.} The primary measure is the node count of the description's canonical abstract syntax tree,
\[
 C_{\mathrm{AST}}(G)=N_{\mathrm{structure}}(d_G)+N_{\mathrm{expression}}(d_G),
\]
where $d_G$ describes game $G$. $N_{\mathrm{structure}}$ counts declaration,
block, and statement nodes; $N_{\mathrm{expression}}$ counts calls, names,
attributes, literals, collections, subscripts, comparisons, and operators.
We exclude parser-only wrapper and context nodes and count each named rule
once at its definition and once at every call site.

\paragraph{RA.} The secondary measure, rule atoms (RA), counts independently changeable rule propositions. Each declaration member or rule statement expresses exactly one such proposition; two effects written in one sentence still count as two. RA is the size of the disjoint union of seven proposition classes: state, action, observation, setup, condition, transition, and outcome.

Both measures are invariant to identifier renaming, comments, and formatting. They depend on the chosen language and measure the size of the rule description, not the state-space size or the difficulty of optimal play.

\subsection{Reading the trajectory plots}
\label{app:trajectory-plots}

The single-run plots present three complementary views of policy development.
Panel (a) plots each submitted snapshot's full-pool Elo, with aligned rank
annotations and frozen-ladder reference levels. These curves can regress;
the row heading separately reports the retained champion.

Panel (b) shows interaction order on 144 equally spaced budget slots
(128 small-match units and 16 full-pool evaluations).
An opponent test or full-pool evaluation occupies one slot.
The continued runs use 432 slots (384 small and 48 full).
Tests in the same batch share the midpoint of its occupied slots.
Small dots give selected opponents' ladder ranks: green denotes wins,
magenta losses, and grey draws. Large diamonds give the learner's full-pool
rank, with rank 1 at the top.
In \textit{Dorado}, 128 opponent tests comprise 16 batches and 256 role outcomes.

Panel (c) reports tokens in millions at each full-pool evaluation.
Lines show cumulative use; bars count tokens since the preceding evaluation,
with the first bar starting at the run's beginning.
Continuation plots annotate ranks at every third full-pool evaluation and at local Elo peaks.

The paired replay-source plots use blue circles for own-policy replays and
purple diamonds for other-policy replays.
Their panel (a) shows both Elo curves and rank rows; dashed levels refer only
to the other-policy run's first evaluated policy and retained champion.
Panel (b) shows cumulative tokens as lines and between-evaluation increments
as paired bars on the right axis.
The ablation bars label retained-champion Elo and rank; colours distinguish
conditions, and panels within each figure share a zero-based Elo axis.

\subsection{Complete Policy Evolution Trajectories}
\label{app:main-trajectories}

These plots show the complete trajectories of the twelve GLM-5.3 runs from
Section~\ref{sec:case-study}.

\begin{figure}[!htbp]
\centering
\begin{figgroup}{grpsuccess}{Success}
\includegraphics[width=\linewidth]{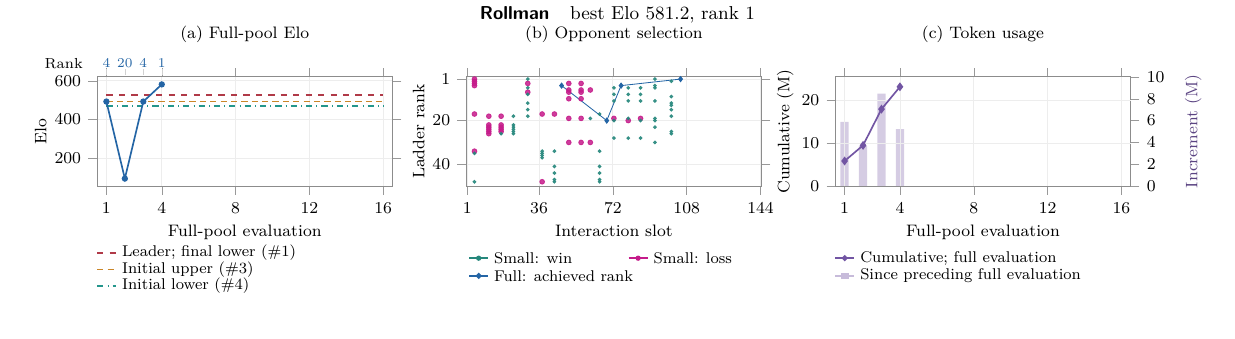}\par\smallskip
\includegraphics[width=\linewidth]{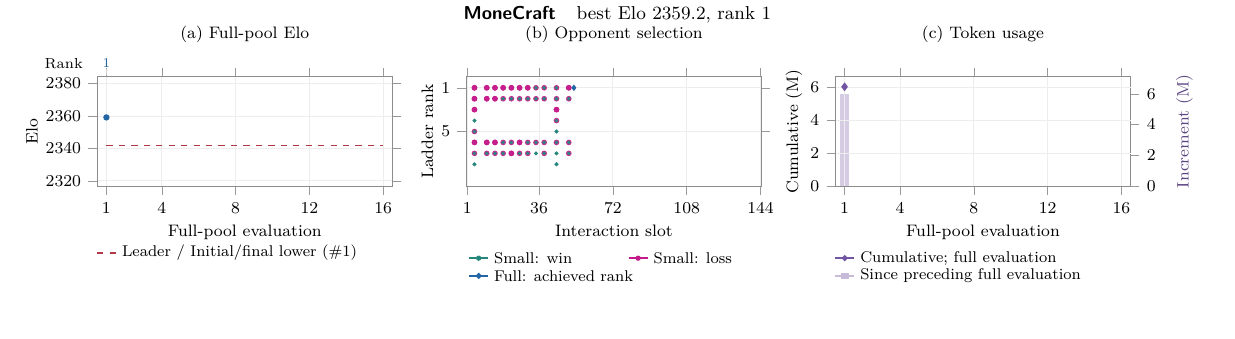}\par\smallskip
\includegraphics[width=\linewidth]{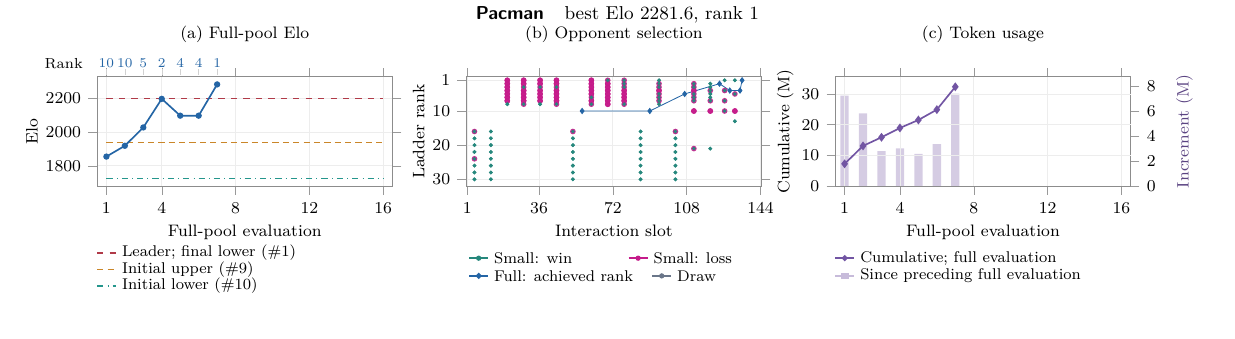}\par\smallskip
\includegraphics[width=\linewidth]{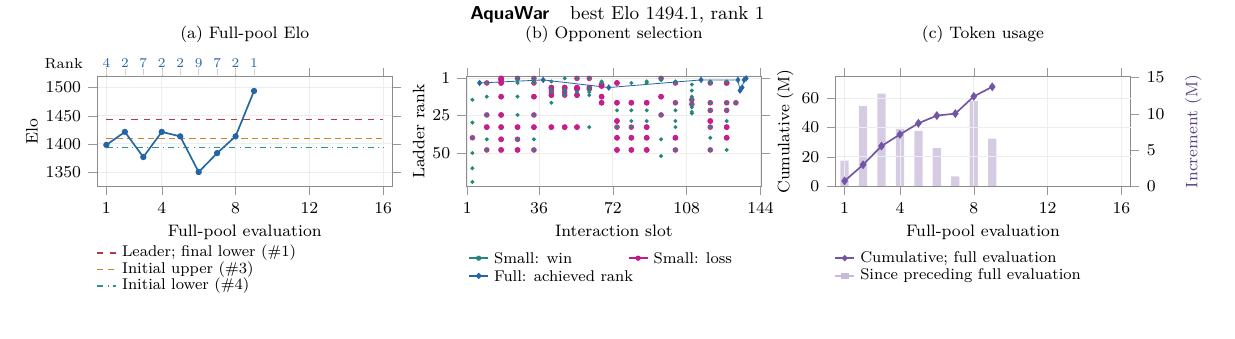}\par\smallskip
\end{figgroup}
\caption{\textbf{Success.} Full-pool Elo and ranks, interaction order, and token use for runs that reach rank 1 and stop early. Plot conventions appear in Appendix~\ref{app:trajectory-plots}.}\label{fig:patterns-early}
\end{figure}

\begin{figure}[!htbp]
\centering
\begin{figgroup}{grpprogress}{Progress}
\includegraphics[width=\linewidth]{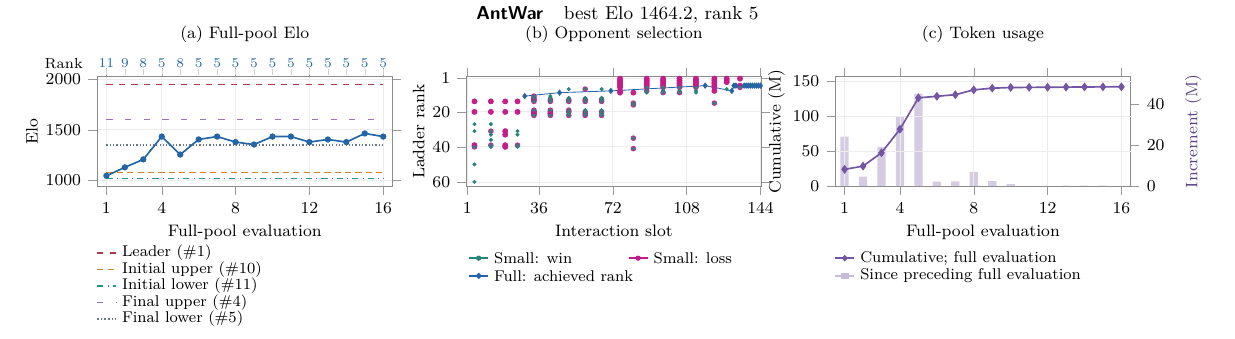}\par\smallskip
\includegraphics[width=\linewidth]{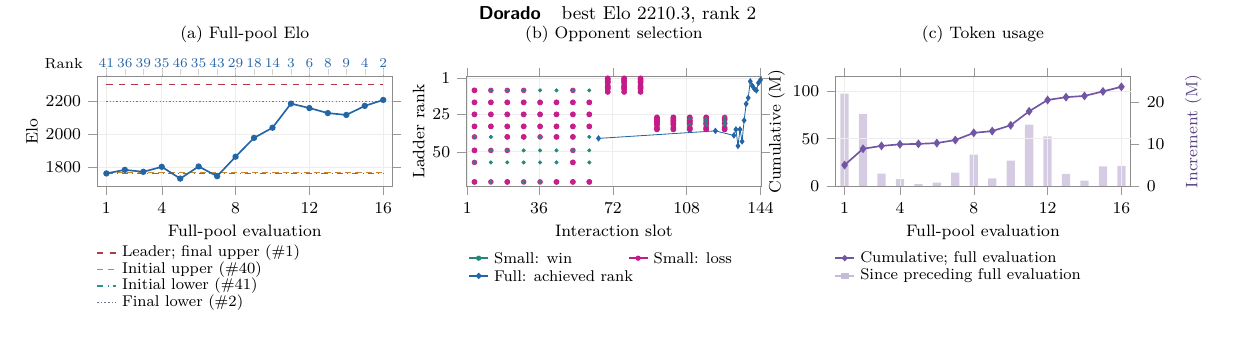}\par\smallskip
\includegraphics[width=\linewidth]{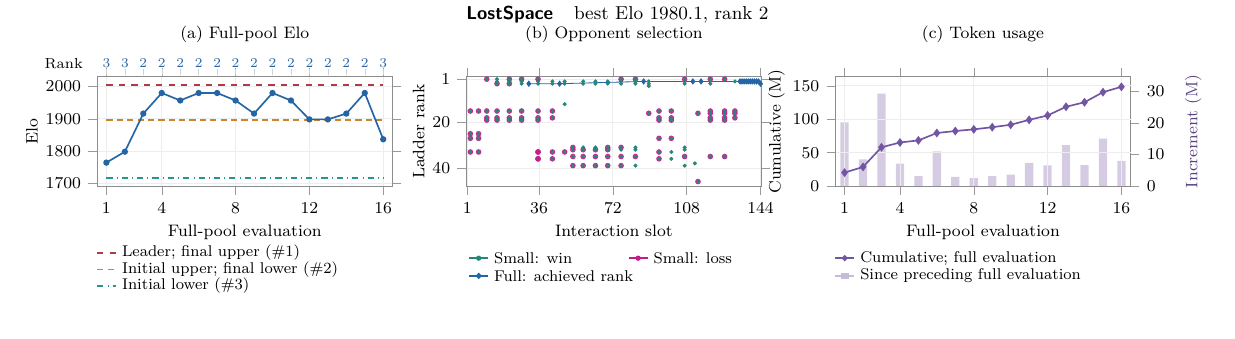}\par\smallskip
\includegraphics[width=\linewidth]{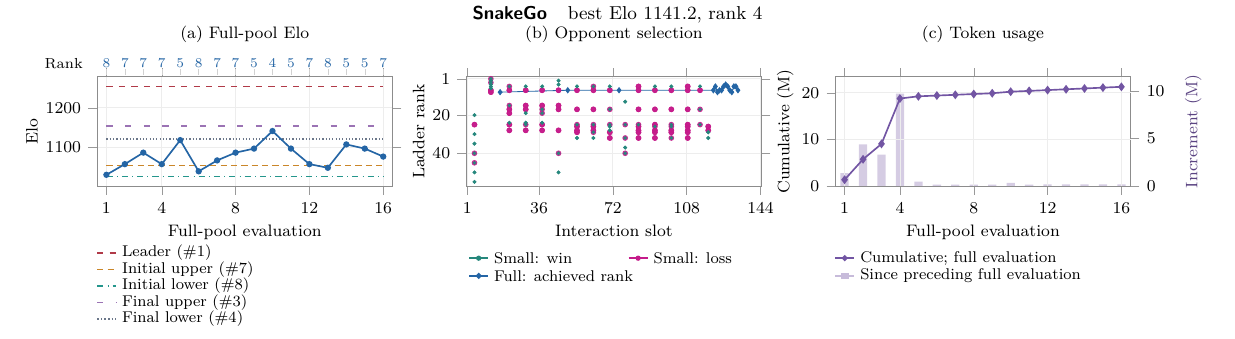}\par\smallskip
\includegraphics[width=\linewidth]{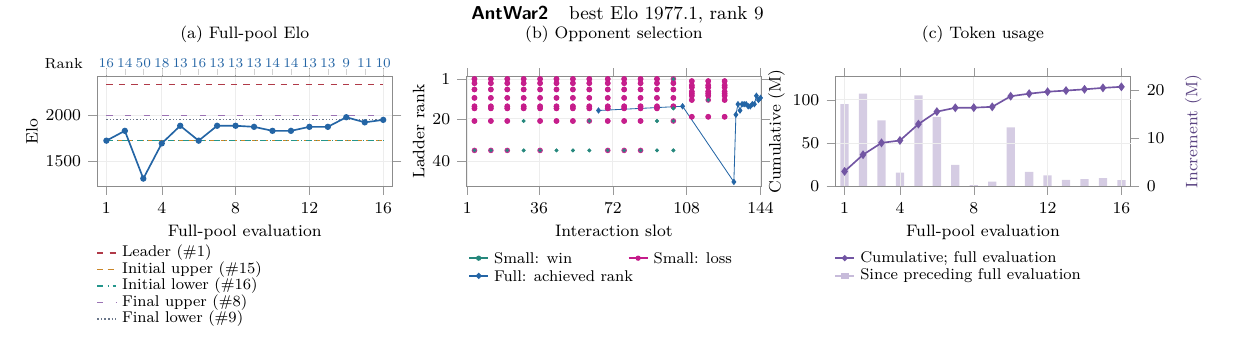}\par\smallskip
\end{figgroup}
\caption{\textbf{Progress.} Full-pool Elo and ranks, interaction order, and token use for games with fluctuating improvement. Plot conventions appear in Appendix~\ref{app:trajectory-plots}.}\label{fig:patterns-rising}
\end{figure}

\begin{figure}[!htbp]
\centering
\begin{figgroup}{grpstagnation}{Stagnation}
\includegraphics[width=\linewidth]{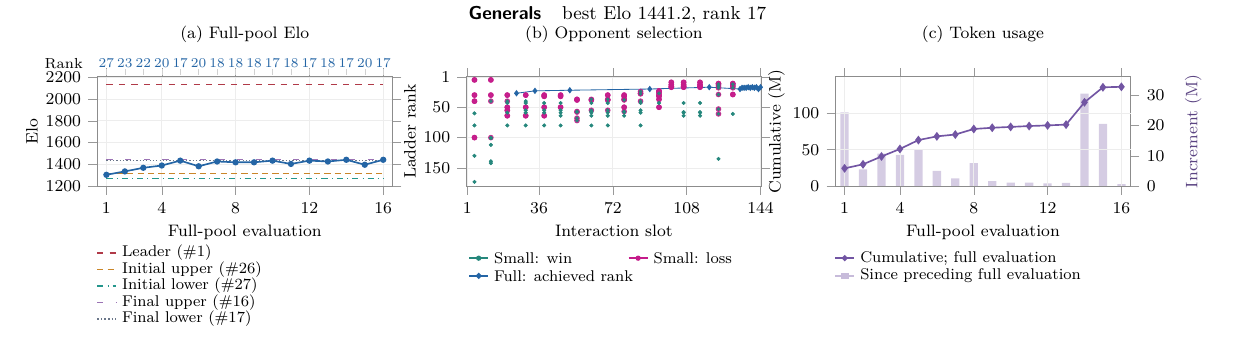}\par\smallskip
\includegraphics[width=\linewidth]{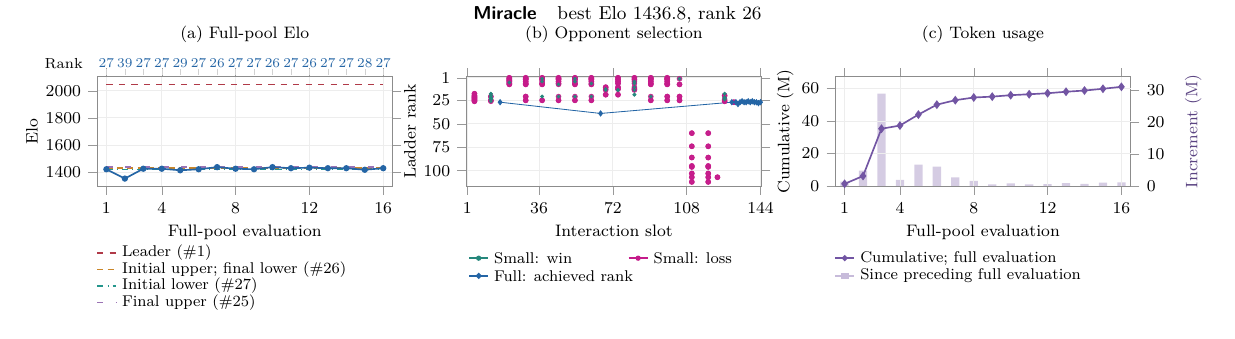}\par\smallskip
\includegraphics[width=\linewidth]{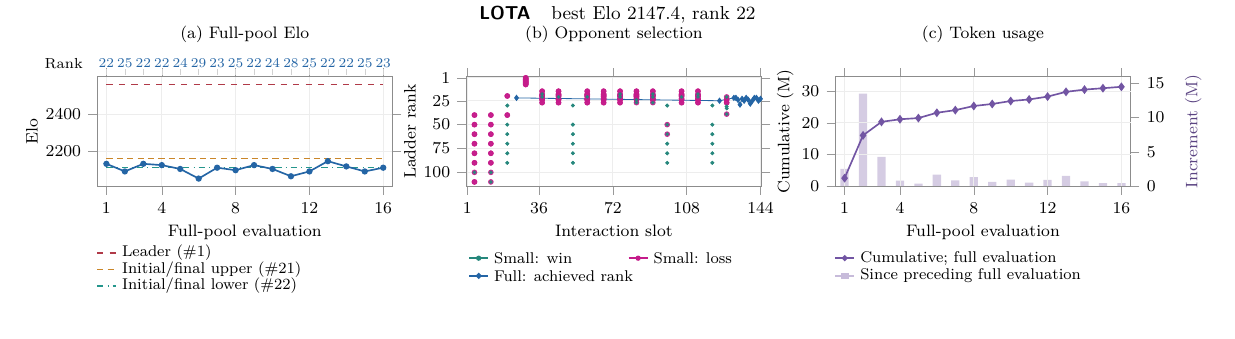}\par\smallskip
\end{figgroup}
\caption{\textbf{Stagnation.} Full-pool Elo and ranks, interaction order, and token use for games that plateau after initial adaptation. Plot conventions appear in Appendix~\ref{app:trajectory-plots}.}\label{fig:patterns-stagnant}
\end{figure}

\FloatBarrier

\section{Detailed ablation results}
\label{app:ablation-details}

Tables~\ref{tab:opponent-ablation}--\ref{tab:batch-ablation} list the Elo and rank behind Figures~\ref{fig:opponent-ablation}--\ref{fig:batch-ablation}.

\begin{table}[!htbp]\centering\small
\begin{tabular}{lrrrr}\toprule
Game & Model & Ladder & Random & Top-4 \\ \midrule
Pacman & 2096.8 / 4 & 2127.5 / 3 & 2014.4 / 6 & 1897.0 / 10 \\
AntWar & 1180.5 / 8 & 1311.7 / 6 & 1154.4 / 9 & 1056.2 / 11 \\
Miracle & 1482.4 / 20 & 1448.8 / 24 & 1457.0 / 24 & 1406.0 / 31 \\
\bottomrule\end{tabular}
\caption{Opponent selection ablation with GLM-5.3. Each cell is Elo / rank under the 128/16 budget.}\label{tab:opponent-ablation}\end{table}

\begin{table}[!htbp]\centering\small
\begin{tabular}{lrr}\toprule
Game & Dense & Binary \\ \midrule
Pacman & 2236.1 / 1 & 1710.1 / 11 \\
AntWar & 1118.7 / 9 & 880.6 / 22 \\
Miracle & 1528.2 / 15 & 1461.1 / 23 \\
\bottomrule\end{tabular}
\caption{Replay feedback ablation with GLM-5.3. Each cell is Elo / rank under the 128/16 budget.}\label{tab:history-ablation}\end{table}

\begin{table}[!htbp]\centering\small
\begin{tabular}{lrrrr}\toprule
Game & 1 & 2 & 4 & 8 \\ \midrule
Pacman & 2040.6 / 5 & 2196.3 / 2 & 2703.9 / 1 & 2160.4 / 2 \\
AntWar & 1130.2 / 9 & 1433.4 / 5 & 1208.7 / 8 & 1118.7 / 9 \\
Miracle & 1338.7 / 45 & 1491.2 / 19 & 1518.7 / 16 & 1465.3 / 23 \\
\bottomrule\end{tabular}
\caption{Small-match batch size ablation with GLM-5.3. Each cell is Elo / rank under the 128/16 budget.}\label{tab:batch-ablation}\end{table}

\section{Token consumption}
\label{app:usage}

The main text summarizes retained-champion performance. Under the same small-match and full-pool limits, the models consume different numbers of tokens. Table~\ref{tab:usage} reports the token use of every main-table run.

\begin{table}[tbp]
\centering\small
\setlength{\tabcolsep}{5pt}
\begin{tabular}{lrrrrrrr}
\toprule
Game & Opus5.5 & GPT6-sol & GLM-5.3 & Kimi K3 & DeepSeek V4 Pro & Qwen 3.8 & LongCat 2.0 \\
\midrule
Rollman & 48.75 & 25.28 & 23.82 & 5.50 & 6.11 & 14.68 & 18.35 \\
Pacman & 22.16 & 16.14 & 33.14 & 40.17 & 91.95 & 163.12 & 17.52 \\
AntWar & 74.19 & 16.08 & 142.94 & 54.59 & 102.80 & 284.29 & 37.56 \\
AquaWar & 76.95 & 26.37 & 68.68 & 1.49 & 60.11 & 13.92 & 20.65 \\
Generals & 75.44 & 19.66 & 137.23 & 88.88 & 172.70 & 60.95 & 22.30 \\
LostSpace & 65.37 & 21.58 & 149.86 & 72.03 & 141.51 & 190.16 & 100.95 \\
Miracle & 89.29 & 19.15 & 61.99 & 34.78 & 47.89 & 107.55 & 49.05 \\
Dorado & 44.58 & 23.02 & 105.63 & 60.72 & 67.01 & 253.02 & 10.14 \\
MoneCraft & 56.16 & 22.10 & 6.75 & 29.97 & 67.54 & 48.72 & 31.18 \\
LOTA & 76.41 & 17.36 & 32.55 & 30.74 & 34.04 & 112.36 & 21.92 \\
SnakeGo & 48.53 & 23.53 & 22.49 & 59.75 & 29.79 & 78.77 & 27.35 \\
AntWar2 & 58.78 & 30.07 & 115.70 & 71.08 & 44.08 & 126.53 & 67.20 \\
\midrule
Total & 736.61 & 260.34 & 900.78 & 549.70 & 865.53 & 1454.07 & 424.17 \\
\bottomrule
\end{tabular}
\caption{\textbf{Token use of the main-table runs} (millions, input plus output; we count cached input once within input). All runs share the 128/16 interaction budget.}
\label{tab:usage}
\end{table}

Equal interaction budgets permit different token expenditure.
Total use varies by more than fivefold: GPT6-sol consumes 260M tokens,
between 16M and 30M per game, whereas Qwen 3.8 consumes 1,454M.
Qwen 3.8 uses 1.6 times GLM-5.3's tokens but earns three gold medals versus four.
GPT6-sol also earns four with less than a third of GLM-5.3's tokens.
The leading configuration, Opus5.5 with Claude Code, consumes 737M.

Token use also varies within a configuration.
Kimi K3 consumes 1.5M on \textit{AquaWar} and 89M on \textit{Generals}.
The feedback ceilings therefore do not impose equal computational expenditure;
we report token use alongside competitive performance.

\section{Algorithm families in the champion policies}
\label{app:code-space}

Because the agent acts entirely in code space, the final policies also show which algorithms each model chooses to write. We classify the code of all 84 retained champions into six families:
\begin{itemize}
\item \textbf{Rule-based control:} hand-written \texttt{if}--\texttt{else} priorities, state machines, and fixed openings or scripts.
\item \textbf{Heuristic scoring:} hand-weighted features that score candidate actions or targets, followed by a greedy choice.
\item \textbf{Graph search:} BFS, DFS, Dijkstra, A*, distance maps, flood fill, and territory partitions.
\item \textbf{Lookahead search:} search over a forward model, including simulated rollouts, beam search, minimax, expectimax, and MCTS.
\item \textbf{Combinatorial optimization:} assignment and matching, dynamic programming, and knapsack-style allocation.
\item \textbf{Inference and randomization:} belief tracking over hidden information, opponent modelling, and mixed or randomized choices.
\end{itemize}

We analyse source files in each champion's strategy directory, excluding
byte-identical SDK files and paths matching version, backup, debugging, test,
or vendor filters. We assign each function a family based on its structure and identifiers.
For functions longer than 40 lines, we classify 20-line windows separately.
We exclude I/O, parsing, and protocol code as infrastructure.
A family's share is its fraction of the remaining non-comment lines.
Table~\ref{tab:code-space} reports each model--game pair and equally weighted
means across models and games.

\input{tables_code-space}

Retained champions rely mainly on rules and heuristic scoring.
Averaging the 84 champion-level shares equally, these families account for 79\%,
graph search for 13\%, and the other three families for 8\%.
Lookahead search exceeds 10\% in only 7 of the 84 champions, all on
\textit{AntWar}, \textit{AntWar2}, and \textit{Generals}.
These endpoint classifications describe policy composition; they do not identify
which code changes caused improvements during development.

Algorithm shares differ across games.
Graph search accounts for 33\% on \textit{Pacman} and 28\% on \textit{Rollman};
inference accounts for 30\% on \textit{AquaWar}.
\textit{MoneCraft} and \textit{LOTA} consist almost entirely of rules.
Across models, Opus5.5 has the lowest rule share (47\%) and the highest
scoring (27\%) and lookahead shares (5\%); GPT6-sol has the highest graph-search share (18\%).
The other models range from 58\% to 65\% rules.

Within each game, we correlate non-rule share ($1-$ rule-based fraction) with Elo
across the seven configurations, using average ranks for ties.
The mean Spearman correlation across 12 games is $0.35$, with variation across games.
This association does not establish that replacing rules improves a policy.

The classification is automatic and approximate.
Identifiers and comments influence it; we assign a mixed function to its dominant mechanism.
Models retaining the same starter code have similar shares, as on
\textit{AquaWar} and \textit{LOTA}.

\end{document}